 \documentclass[final,5p,times,twocolumn]{elsarticle}

\usepackage{amssymb}
\usepackage{graphicx} 
\usepackage{subcaption} 
\usepackage{float}
\usepackage{natbib}
\usepackage{hyperref}
\usepackage{tabularx}
\usepackage{xcolor}
\usepackage{graphicx}
\usepackage{caption}
\usepackage{subcaption}
\usepackage{booktabs}
\usepackage[utf8]{inputenc}
\usepackage{geometry}

\journal{UNSW}

\begin{document}

\begin{frontmatter}



\title{Large language models for sentiment analysis of  newspaper articles during COVID-19: The Guardian}


            \author[inst1]{Rohitash Chandra}
\author[inst1]{Baicheng Zhu} 
\author[inst1]{Qingying Fang} 
\author[inst1]{Eka Shinjikashvili}
\affiliation[inst1]{Transitional Artificial Intelligence Research Group, School of Mathematics and Statistic, UNSW, Sydney, Australia }




\begin{abstract}
During the COVID-19 pandemic, the news media coverage encompassed a wide range of topics that includes viral transmission, allocation of medical resources, and government response measures. There have been studies on sentiment analysis of social media platforms during COVID-19 to understand the public response given the rise of cases and government strategies implemented to control the spread of the virus.  Sentiment analysis can provide a better understanding of changes in societal opinions and emotional trends during the pandemic.
Apart from social media, newspapers have played a vital role in the dissemination of information, including information from the government, experts, and also the public about various topics. A study of sentiment analysis of newspaper sources during COVID-19 for selected countries can give an overview of how the media covered the pandemic. 
In this study, we select The Guardian newspaper and provide a sentiment analysis during various stages of COVID-19 that includes initial transmission, lockdowns and vaccination.  We employ novel large language models (LLMs) and refine them with expert-labelled sentiment analysis data.  We also provide an analysis of sentiments experienced pre-pandemic for comparison.  The results indicate that during the early pandemic stages, public sentiment prioritised urgent crisis response, later shifting focus to addressing the impact on health and the economy.  In comparison with related studies about social media sentiment analyses, we found a discrepancy between The Guardian with dominance of negative sentiments (sad, annoyed, anxious and denial), suggesting that social media offers a more diversified emotional reflection. We found a grim narrative in The Guardian with overall dominance of negative sentiments, pre and during COVID-19 across news sections including  Australia, UK, World News, and Opinion.

\end{abstract}




\end{frontmatter}

\section{Introduction}
\label{sec:sample1}
The COVID-19 pandemic was a highly contagious \cite{2020Coronavirus46} and catastrophic global event with infection symptoms such as fever and cough, with severe cases potentially leading to death \cite{koh2021deaths}. Various measures were implemented to control the spread of COVID-19 such as patient isolation, mandatory masks in public places, social distancing, and virus testing and vaccination \cite{guner2020covid, 2023Vaccine, jin2020virology}. The pandemic led to social changes, including lockdowns, quarantines, economic downturns and loss of life, which have affected mental health \cite{2021CovidStress}. Additionally, social uncertainty and fear brought about by the COVID-19 pandemic have led to social unrest and instability \cite{2021CovidUnrest}. During the COVID-19 pandemic, major social media platforms such as Facebook, Weibo, and Twitter were widely used to disseminate information about infection cases \cite{sahni2020role}. They included government safety and control measures, vaccine and drug development, and various other aspects related to the pandemic \cite{goel2020social}.  Social media users of these platforms also shared their thoughts on the pandemic and expressed various concerns and sentiments \cite{chandra2021covid}. Social media has also been used as a medium for misinformation during COVID-19 \cite{tsao2021social} that resulted in measures by social media such as the removal of fake accounts and better management of content \cite{ahmed2022social}.

Natural Language Processing (NLP) \cite{chowdhary2020natural} is a field of artificial intelligence that enables computers to understand, interpret, manipulate, and generate human language \cite{chowdhary2020natural}. The development of NLP has benefited from advances in machine learning and deep learning models\cite{otter2020survey, wu2020deep, li2018deep},  enabling NLP systems to yield favourable outcomes when tackling intricate language tasks \cite{jones1994natural}. Large Language Models (LLMs) refer to neural network models with a large number of parameters and extensive pre-training, primarily employed for natural language understanding tasks \cite{brown2020language}. Sentiment analysis \cite{medhat2014sentiment} in NLP aims to identify and understand the emotional tendencies or sentiments expressed within a text, typically by detecting sentiments such as positive, negative, and neutral \cite{taboada2016sentiment}. Sentiment analysis has extensive applications across various domains, including social media analysis, public sentiment monitoring, product review analysis, and customer feedback analysis \cite{medhat2014sentiment}. It has been demonstrated to be used not only for social media and advertising industry \cite{medhat2014sentiment, wankhade2022survey, sanchez2020opinion}, but can also be used to study public behaviour in pandemic such as the COVID-19 \cite{chandra2021covid} and compare texts to evaluate the quality of translations from low-resource languages \cite{shukla2023evaluation}.

Recurrent Neural Networks (RNNs) \cite{goodfellow2016deep, lipton2015critical} are deep learning models for processing sequential data and are prominent for processing text data and performing tasks such as sentiment analysis. The Long Short-Term Memory (LSTM) network \cite{hochreiter1997long, yu2019review} is an advanced RNN for modelling data that have long-term dependencies. In the last decade, variants of the LSTM network have emerged, and the encoder-decoder LSTM \cite{malhotra2016lstm} became a prominent variant for processing and modelling text data.  The Transformer model is an advanced encoder-decoder LSTM with attention mechanism \cite{vaswani2017attention} that was designed for NLP tasks, particularly for tasks involving sequential data. It is currently the mainstream framework for building LLMs. The Transformer model lays the foundation of the BERT (Bidirectional Encoder Representations from Transformers) \cite{devlin2019bert} model that is a pre-trained LLM used for various NLP tasks, including sentiment analysis \cite{acheampong2021transformer}. RoBERTa (Robustly optimised BERT approach) \cite{liu2019roberta} is a variant of BERT that utilises larger-scale datasets for pre-training, longer sequence lengths, and dynamic masking. These improvements enhanced the RoBERTa model performance and generalisation capability for sentiment analysis. BERT-based LLMs have been widely applied in sentiment analysis of content on social media. Wahyudi and Sibaroni \cite{2022Deep} used BERT for sentiment analysis of TikTok  based on the reviews obtained from the Google Play Store and the Apple App Store. Kikkisetti et al. \cite{kikkisetti2024using} utilised LLMs for identifying emerging coded antisemitic hate speech on extremist social media platforms. 
 
During the COVID-19 pandemic, major news media outlets continued to track and report on the development of the epidemic \cite{hart2020politicization}, such as the US Daily newspaper \cite{basch2020coverage}, China Daily \cite{yu2021corpus},  and The Guardian \cite{yu2021corpus}. Due to the large volume of news articles, researchers employed NLP methods to analyse news.  Some of the prominent NLP applications for COVID-19 news are as follows. Kocaman et al. \cite{kocaman2021spark}  used the clinical and biomedical NLP models from the \textit{Spark NLP} \footnote{\url{https://sparknlp.org/}} for healthcare library to analyze COVID-19 news publications from the \textit{CNN} and  \textit{Guardian} newspapers \cite{varol2022understanding}. Evans and Jones \cite{evans2023emotional}  used NLP to analyze news from major UK news channels, aiming to explore the widespread negative impact of the COVID-19 pandemic on mental health. Through NLP, we can gain insights into the public opinion trends and the epidemic situation of COVID-19 across different countries. For example, Tejedor et al. \cite{tejedor2020information} reported that  Spain and Italy exhibited a high degree of politicisation in epidemic control, and  Apuke et al. \cite{apuke2020nigerian} reported that the news media in Nigeria focused more on reporting death tolls and typical cases rather than epidemic prevention measures, which led to public panic. These motivate further studies of newspaper reporting during COVID-19 via NLP methods such as sentiment analysis.

The Guardian is a prominent international newspaper that covers COVID-19, hence we discuss some of the relevant studies in the analyses of its news reporting.  
Sheshadri et al. \cite{sheshadri2017no} conducted an analysis of selected news articles from The Guardian and The New York Times to investigate how certain hierarchies are expressed and evolve across time and place. The authors found that media portrayals of religious groups remained relatively consistent, showing stability across decades rather than being significantly influenced by immediate events. Tunca et al. \cite{tunca2023exploratory} employed NLP methods to analyze news articles from The Guardian to identify sentiment characteristics and reveal the interrelationships between the concept of the metaverse \cite{mystakidis2022metaverse} and other relevant concepts. The study reported that positive discourses about the metaverse were associated with key innovations for users and companies, and negative discourses were associated with issues relating to the use of social media platforms such as privacy, security and abuse. Abbasian et al. \cite{abbasian2017uk} focused on media reflection during the Iraq War to examine the portrayal of British Prime Minister Tony Blair in The Guardian, and observed that Blair was portrayed as a war criminal by this newspaper.

Sentiment analysis has been used to identify mental health problems during COVID-19 for psychological intervention \cite{Yousefi2021isolation}. Sentiment analysis can be used to evaluate the effect of policy measures, since different policy measures may have different sentiments and reactions to the public. Goel et al. \cite{goel2022investor} assessed the impact of various policies on public sentiments,   which provided feedback for government decision-making. Through sentiment analysis, we can understand the public's reaction to information during COVID-19, such as the COVID-19 vaccines \cite{info:doi/10.2196/30765}. By identifying positive and negative sentiments, we can determine the overall public response to vaccines and news events related to vaccines, to accurately convey information and guide the public to face the challenge with a positive attitude \cite{info:doi/10.2196/30765}. Therefore, sentiment analysis can help us better understand the public's emotional changes before and during COVID-19. 
 
In this study, we analyze the sentiments expressed in The Guardian newspaper before and during the COVID-19 pandemic. We investigate selected article sections and compare the difference in sentiment expressions between the World, UK and Australian article sections.  We present a framework that employs pre-trained LLMs that include  BERT and the RoBERTa models.   The main goal of our study is to provide sentiment analysis prior to, and during COVID-19.  This can provide guidance for policymaking, crisis response, psychological intervention and public opinion guidance in planning for the management of critical events. We compare the sentiments obtained during COVID-19 with the trend of COVID-19 infections (death) and also provide analysis via a quarterly timeframe.

The rest of the study is organised as follows. In Section 2, we provide a background on deep learning models for LLMs. Section 3 presents the methodology, including data processing and modelling. In Section 4, we present and discuss the results and Section 5 provides a discussion of the major results and limitations.  Finally, we conclude the paper in Section 6.
 
\section{Background}
\label{sec:back}

\subsection{Recurrent Neural Networks}

Unlike traditional feedforward neural networks, RNNs possess recurrent connections, enabling them to capture temporal dependencies within sequential data \cite{Elman1990, 2014Deep}.  RNN has significant advantages in handling sequential data and capturing temporal dependencies. The recurrent connections in RNN enable it to capture temporal dependencies within the data, thereby better understanding the contextual information of the sequences making them applicable to NLP \cite{2015A}. Also, the recurrent structure enables the model to memorise information from previous time steps and propagate it into future steps. In terms of data input, RNN has a certain degree of flexibility and can handle input sequences of various lengths.  
 
LSTM networks \cite{yu2019review} provide an extension to simple RNNs for modelling long-term dependencies in sequence data \cite{hochreiter1997long}, making them applicable to NLP problems. The networks address the issue of vanishing gradients \cite{1998The} encountered by simple RNNs.   LSTM networks require substantial training time and computational resources, given the complex structure. LSTMs demand significant computational power and memory, especially when training on large datasets \cite{2015Critical}. Additionally, LSTMs are susceptible to overfitting, particularly in scenarios with relatively small data volumes, although this issue can be mitigated through gularization methods such as  dropouts \cite{2014Dropout}. Despite these limitations, due to their ability to capture long-term dependencies and generalization performance, LSTMs remain an effective tool for NLP.

\subsection{Transformer model}

The Transformer model is an advanced RNN effective in processing sequential data, such as text or time series information \cite{Attention2017}.  
The Transformer model  utilises an \textit{attention} motivated by human behaviour and utilises the bidirectional LSTM network \cite{BiLSTM2005}.   The attentional mechanism is crucial for understanding the key points in sentences or paragraphs and provides better contextual understanding.   Training on large datasets also aids the model in understanding rare words or expressions \cite{LuWei2022RooB}. Additionally, the Transformer model can be fine-tuned for specific tasks and the accuracy can be improved based on large-scale datasets for specific tasks \cite{bansal2023adaptation}.
However, particularly large Transformer models require significant computational resources for training, including considerable time and memory resources, which makes the cost of maintaining these models quite high \cite{CunhaWashington2023ACSo}. Furthermore, fine-tuning Transformer models on small datasets may lead to overfitting, where the model excessively learns specific patterns in the training data at the expense of its generalisation ability \cite{LiGuangju2024MTwm}. 

\subsection{BERT and RoBERTa models}

The BERT is a pre-trained LLM based on the Transformer model
 \cite{devlin2019bert} and capable of understanding the bidirectional context of each word within the text. This is distinctly different from conventional unidirectional or fixed-direction language models.  
 BERT operates in two main phases: pre-training and fine-tuning. During pre-training, it engages in two types of self-supervised learning: masked language modelling (MLM), which involves predicting tokens for randomly masked positions in the input, and next sentence prediction (NSP), which determines if two segments of text are sequentially related. For fine-tuning, which tailors the model for specific tasks, additional fully connected layers are appended above the final encoder output. The inputs can be prepared by tokenising text into subwords, then merging three types of embeddings - token, positional, and segment - to form a standardised vector. 

The RoBERTa model \cite{liu2019roberta}  is a pre-trained LLM that enhances performance and robustness based on BERT. It adopts the "robust optimization BERT method" in its design. In comparison to the original BERT model, RoBERTa shows several advantages and drawbacks.
RoBERTa benefits from a larger training corpus, containing extensive text data from web pages, forums, books, and more, thereby enhancing the model's performance.
The model employs a stricter masking strategy (dynamic masking) by replacing all words in the input text with "[MASK]", facilitating more effective utilization of training data information.
Furthermore, RoBERTa adopts a deeper network architecture, which contributes to further improving the model's performance. 
Finally, it demonstrates superior performance across various natural language processing tasks, including the GLUE (General Language Understanding Evaluation) \cite{wang2019glue} and SuperGLUE \cite{wang2020superglue} task sets.
However, there are drawbacks to consider since RoBERTa requires a larger volume of text data and a deeper network structure in comparison to BERT, resulting in longer training times.
The substantial model size of RoBERTa poses challenges in terms of resource requirements during deployment.


\section{Methodology}

\subsection{Data}

\subsubsection{SenWave sentiment dataset}
The SenWave dataset \cite{yang2020senwave} aggregated over 105 million tweets and Weibo messages on COVID-19 in six distinct languages: English, Spanish, French, Arabic, Italian, and Chinese, from March 1 to May 15, 2020. This dataset features 10,000 English and 10,000 Arabic tweets that have been tagged across 10 specific sentiment categories. Additionally, it includes 21,173 Weibo posts categorised into 7 different sentiment types for the Mandarin language. In total, the dataset contains 41,000 labelled items and over 105 million unlabelled tweets and Weibo posts, positioning it as the most extensive labelled COVID-19 sentiment analysis dataset. In our study, we utilise a subset of the SenWave dataset\footnote{SenWave dataset: \url{https://github.com/gitdevqiang/SenWave/blob/main/labeledtweets/labeledEn.csv}}, specifically focusing on 10,000 English tweets with sentiment labels such as: optimistic, thankful, empathetic, pessimistic, anxious, sad, annoyed, denial, official COVID-19 report, and joking. 

The first phase of our framework focuses on the preprocessing of SenWave data using  NLP  tools that include converting text to lowercase, expanding contractions, translating emojis to text, and stripping special characters as well as URLs using regular expressions and custom functions as done by Chandra and Krishna \cite{chandra2021covid}. This process ensures that the data is homogenised and primed for further analysis. 
We train our model using the SenWave dataset, and then apply sentiment analysis on The Guardian newspaper articles covering COVID-19.



\subsubsection{The Guardian newspaper}
The Guardian is a well-known British news media which covers local and international news. 
\textit{Kaggle}\footnote{\url{https://www.kaggle.com/}} is a popular platform for the data science community, where people can find a rich variety of datasets covering fields such as finance, healthcare, image processing,  and NLP. Kaggle has been prominent in running forecasting and related machine learning competitions \cite{kaggleforecasting}. The Guardian newspaper dataset on Kaggle \cite{kaggle_dataset} comprises approximately 150,000 news articles from 2016 to 2023. We extracted and partitioned the dataset into two main intervals based on the World Health Organization’s (WHO) COVID-19 timeline \cite{worldhealthorganization_2021_timeline}: pre-pandemic (January 1, 2018, to December 31, 2019) and during the pandemic (January 1, 2020, to March 31, 2022). We focus on quarterly  (3 months) analysis of the data, and divided the dataset into nine distinct periods for observing the evolution of media sentiments as the pandemic unfolded. This is also to identify shifts in focus and tone within the news, revealing how the crisis influenced reporting priorities and language use over time.

The Guardian dataset is categorised into 164 sections, and a significant number of these sections contains only one or two articles, hence, we excluded these sections from further analysis. We selected only four major news sections for detailed examination, including World News, Opinion, Australia News, and UK News. We omitted both the Football and Sports sections, despite their substantial data (combined volume exceeding 10,000 articles, as shown in Table \ref{tab:count}). The rationale behind this exclusion lies in the assumption that sports-related content would unlikely contribute meaningful insights into the sentiment analysis relevant to our study.

\begin{table}[htbp!]
\centering
\begin{tabular}{lc}
\toprule
\textbf{Section Name} & \textbf{Article Count} \\
\midrule
World News        & 11269 \\
Opinion           & 7912  \\
Football          & 7135  \\
Sport             & 6633  \\
Australia News    & 5419  \\
US News           & 5004  \\
Business          & 4672  \\
Politics          & 4606  \\
UK News           & 4255  \\
\bottomrule
\end{tabular}
\caption{The Guardian's main sections and article counts.}
\label{tab:count}
\end{table}

Our study primarily concentrates on comparing sentiment trends between the UK and Australian news, acknowledging a discrepancy in the volume of articles, with the Australian dataset surpassing the UK’s by approximately 1,000 articles. Additionally, the Opinion section was specifically selected for its unique editorial stance. Unlike other sections aiming for neutrality, the Opinion section inherently encompasses a broader spectrum of sentiments and viewpoints, providing a rich source for dissecting sentiment shifts before and amidst the pandemic. This choice stems from the premise that opinion pieces, by nature, are more likely to express sentiments that deviate from neutrality, thereby offering invaluable insights into the public and editorial sentiment during these contrasting periods.

\subsection{N-gram analysis}

N-gram is a commonly used for NLP  that reviews contiguous N-elements (usually words or characters) in text data \cite{TRIPATHY2016117}. The N-gram model assumes that each element in the sequence depends only on its previous element, and utilises this local dependency to make predictions or analyze data. The advantages of the N-gram model include simplicity, computational efficiency, and its ability to capture local information \cite{WuHaiyan2021KnEa}. However, the N-gram model also has some limitations, such as its inability to capture long-distance dependencies, the need for large amounts of training data, and challenges in handling rare combinations. The application of N-gram models in news media analysis is extensively documented, reflecting the diversity and practicality of this method in real-world research.  Lyse and Andersen \cite{Lyse2012} used N-gram in news media analysis to rank word sequences from the Norwegian Newspaper Corpus  to assess the propensity co-occurrence of binary and triplets. Furthermore, N-gram-based text analyses were used for selected Persian \cite{2011N}, German  \cite{1981Lemmatizing}, and English newspapers \cite{2016N}. We will use N-grams to provide analysis of the data in association with sentiment analysis by LLMs.


\subsection{Fine-tuning BERT and RoBERTa models}


Given the diverse range of topics and writing styles found in our news articles dataset, fine-tuning the BERT and RoBERTa models are crucial so that they can effectively provide sentiment analysis with the 10 sentiments available in the SenWave training data \cite{yang2020senwave}.
However, fine-tuning BERT also encounters numerous challenges. First of all, the fine-tuning process requires setting up a suitable environment to enhance efficiency and effectiveness \cite{MirandaCarlosHenríquez2023EtEo}. Furthermore, news articles often present lengthy content, which will surpass BERT's token limit. Segmenting these articles and aggregating sentiment scores poses another challenge, necessitating careful attention to maintain context and coherence \cite{MutindaJames2023SAoT}. Moreover, the presence of sarcasm and nuanced sentiments in news articles can perplex the model, which will potentially hinder accurate interpretation. Therefore, handling specialised terminology, proper nouns, or novel words absent from BERT's vocabulary is important for maintaining sentiment analysis accuracy \cite{DangXiaochao2023DRoN}. Therefore, addressing these challenges needs a multifaceted approach involving meticulous data preprocessing, domain-specific fine-tuning, and appropriate parameter setting. These measures collectively ensure the accurate and reliable sentiment analysis of news articles using BERT.

\subsection{Framework}

\begin{figure*}[ht]
    \centering
    \includegraphics[width=1\textwidth]{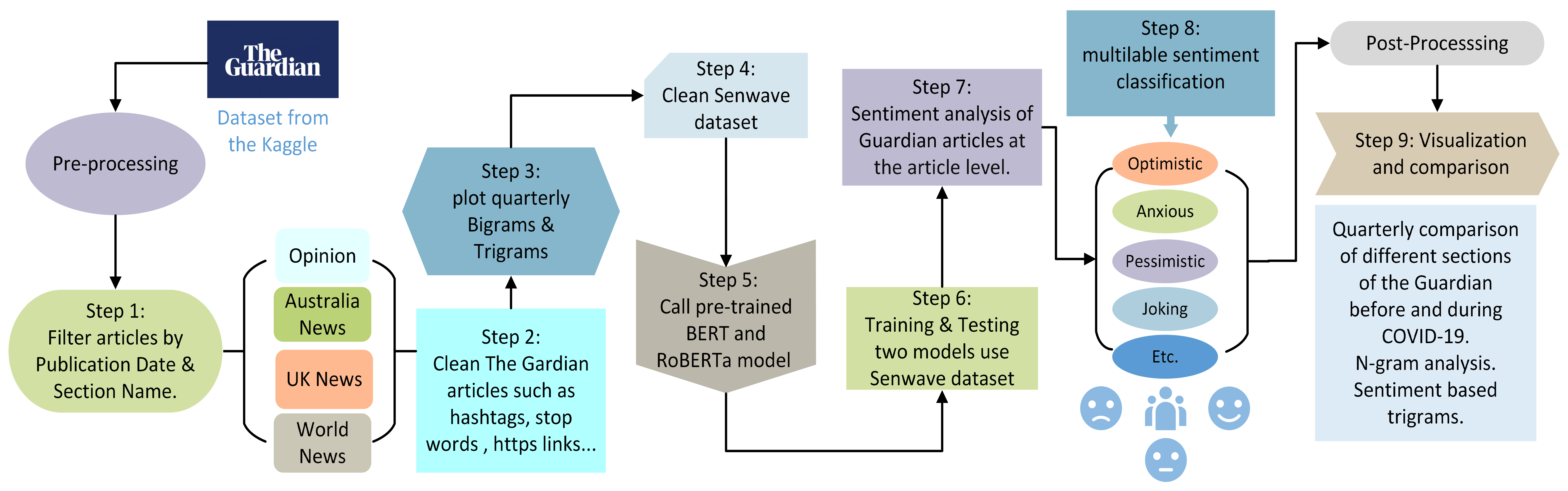}
    \caption{A sentiment analysis framework for The Guardian articles (01/01/2018 - 31/03/2022) utilising BERT and RoBERTa models trained and tested on the Senwave dataset.}

    
    \label{framework}
\end{figure*}


Our framework for using deep learning for sentiment analysis of newspaper articles includes several consecutive steps, as shown in Figure \ref{framework}. The major components include: 1. Guardian data selection; 2. data cleaning; 3. bigram and trigram visualisation; 4. SenWave dataset cleaning; 5. pre-trained model calling; 6. model training and testing; 7. document-level sentiment analysis; 8. multi-label sentiment classification; 9. visualisation and comparison. 

Our framework uses two data sets, including the Guardian articles and the SenWave dataset. We use the SenWave dataset for refining the LLM models. We first obtain the Guardian dataset from Kaggle and filter it based on publication dates and section names. We extract articles tagged (sections) as Australian news, UK news, World news, Business, Politics and Opinion. Note that the Opinion section encompasses local (UK) and international contributors.  We believe that these sections are more important in sentiment analysis during COVID-19 since we cover the country of source of the Guardian, the world, and Australia which is geographically far but politically aligned with Great Britain.

In data processing (Step 2), we remove the items such as hashtags, stop words and web hyperlinks from the Guardian data using NLTK (Natural Language Toolkit) \cite{loper2002nltk} tools.  In order to obtain more meaningful bigrams and trigrams, and subsequent sentiment analysis, we also remove stop words based on categories, such as general terms, as shown in Table \ref{tab:process}. 

In Step 3, we segment the Guardian data into words and character sequences. In each Guardian article, starting from  2018, we generate all possible bigrams and trigrams for each quarter (data spanning every three months).  We then visualise the bigrams and trigrams and analyse word combinations and frequencies to gain insight into article characteristics across different periods (quarters) covering COVID-19.

 In Step 4, we extend our data cleaning process to the SenWave dataset, which consists of tweet texts, unlike the formal media articles from The Guardian. Given the dynamic nature of social language, we must account for abbreviations, emoticons, and other informal elements. To accomplish this, we follow the methodology outlined by Chandra and Krishna \cite{chandra2021covid} for processing the SenWave dataset, ensuring it is ready for subsequent model training and testing. It's worth noting that within the SenWave dataset, the term "official report", while typically considered a topic, is classified as a sentiment. This distinction is noteworthy, especially considering that in our analysis of The Guardian articles, sections discussing "official reports" comprise a significant portion of the dataset. 

 In Step 5, we leverage pre-trained BERT and RoBERTa models via Hugging Face \footnote{Hugging Face: \url{https://huggingface.co/models}}. The Hugging Face is a leading company in the field of artificial intelligence dedicated to providing open-source NLP tools \cite{chhabra2024exploring}. One of their flagship products is the Transformer model library, which provides a variety of pre-trained deep learning models, including BERT and RoBERTa. Hence, we can access these state-of-the-art models and fine-tune them for our sentiment analysis tasks to ensure optimal preparation.

In Step 6, we employ the cleaned SenWave dataset for refining  the pre-trained models (BERT and RoBERTa)   using  GLoVe embedding \cite{pennington2014glove} for multi-label sentiment classification. We note that BERT and RoBERTa can be used without GloVe embeddings, but we used GloVe to follow the work of Chandra and Krishna \cite{chandra2021covid}.




We begin by setting parameters, including maximum token length, batch sizes, epochs, and the learning rate to tailor the training process. Using the BERT and RoBERTa Tokenizers from the Hugging Face Transformers library \cite{huggingface}, we convert the tweets into a compatible format for processing. This format includes the necessary attention masks and token type (given by identity number (IDs)), ensuring the model can effectively learn from our data. We enhance the model with a dropout layer for regularisation and a linear layer for adjusting the output, thus preparing it for robust training sessions. Training involves iterating over batches, where the model computes losses and updates weights accordingly. After training, we evaluate the model's performance using metrics such as Hamming loss, Jaccard score, label ranking average precision (LRAP) score, and F1 scores. These metrics provide critical insights into the model’s effectiveness in handling multi-label classification tasks, underscoring our comprehensive approach from data preparation to model training and evaluation in a deep learning framework.

In Step 7, we utilise the fine-tuned BERT and RoBERTa models to conduct sentiment analysis on selected articles from the Guardian.  

In Step 8, we conduct a multi-label sentiment classification of the Guardian data across the different quarters by categorising sentiments such as optimism, anxiety, pessimism, and surprise. This approach allows for a more nuanced understanding of the emotional nuances expressed within the articles during different phases of COVID-19 that spans quarter before and during the COVID-19 pandemic, covering the beginning, lockdowns, vaccination, end of lockdowns and unrestricted travel, locally and internationally, which marks the end of COVID-19. 

In the final step (Step 9), we employ various visualisation techniques, including bigram and trigram analysis for selected sentiments, sentiment polarity, sentiment distribution histograms, and heatmaps. We compare the different sections of The Guardian before and during the COVID-19 pandemic. Through these visualisations, we aim to illustrate the evolution of sentiments and topics over time.

\begin{table}[htbp]
\small
    \centering
    \begin{tabular}{|>{\raggedright\arraybackslash}p{0.25\linewidth}|>{\raggedright\arraybackslash}p{0.7\linewidth}|}
    \hline
    \textbf{Category} & \textbf{Stop words} \\
    \hline
    General terms & guardian, article, theguardian, com, says, also, world, news, report, way, ways, use, used, using, place, places \\
    \hline
    People and entities & man, men, woman, women, child, children, company, companies, minster, government, scott, morrison, police, officer, boris, johnson, federal, prime, york, zealand \\
    \hline
    Action verbs & make, makes, made, go, going, get, getting, got, see, seeing, seen, take, takes, took, come, comes, came, look, looks, looking, help, want, wants, wanted, tell, tells, told, work, works, worked \\
    \hline
    Time-related & year, years, time, times, week, weeks, month, months, day, days \\
    \hline
    Quantifiers and qualifiers & one, many, much, lot, lots, number, numbers, group, groups, good, bad, important, need, needs, needed, including \\
    \hline
    Modifiers and auxiliaries & said, just, like, can, may, might, think, thinks, thought, should, must, do, does, will, would, could, south, wale \\
    \hline
    \end{tabular}
    \caption{Categorised extra stop words used only for the N-gram analysis. These stop words are not part of the default  English dictionary of stop words obtained from NLTK\protect\footnote{\url{https://www.nltk.org/search.html?q=stopwords}}}.  
    \label{tab:process}
\end{table}

\subsubsection{Technical details}

Our framework in Figure \ref{framework} leveraged the BERT-based pre-trained models from  Hugging Face \cite{huggingface}. We implemented our framework in Python with libraries such as NumPy and Pandas for data manipulation, Matplotlib and Seaborn for data visualisation, and NLTK\footnote{\url{https://www.nltk.org/}} with TextBlob\footnote{\url{https://textblob.readthedocs.io/en/dev/}} for NLP tasks. We also used PyTorch library that supports vector and tensor operations and model training processes, alongside TorchText\footnote{\url{https://pytorch.org/text/stable/index.html}} library for handling text data.  The entire open source code and data is provided in our GitHub repository with details at end of the paper.

Both models (BERT and RoBERTa) were fine-tuned to approximately 400 megabytes in size using SenWave dataset for sentiment analysis. We used the BERT-base-uncased version, which contains approximately 110 million parameters, and RoBERTa-base with 120 million parameters, and employed the BertTokenizer to prepare the input data with a maximum sequence length of 200 tokens to align with the average length of tweets.

 We refined the respective LLMs  in a GPU-accelerated environment to manage the substantial computational load, with each model taking approximately 4 hours for training and fine-tuning.  We used a computer with 1 Intel Core i7-12700H processor   and 16 gigabytes of memory.   In order to validate the SenWave-based refining of the model, we implemented a 90:10 ratio for training and test sets. We employed a relatively low learning rate of 1e-05 for BERT and 2e-05 for RoBERTa, to achieve finer adjustments in the model weights during backpropagation. The training (refinement) process spanned 4 epochs and throughout the training process, the loss of the models showed a decreasing trend, reflecting continuous learning and improvement. The models were trained using a batch size of 8 for both the training and validation phases, which is suitable given the complexity of sentiment classification tasks and the depth of the models.



\section{Results}

\subsection{COVID-19 - cases and deaths }

\begin{figure}[htbp!]
    \centering
    \begin{subfigure}[b]{0.5\textwidth}
        \centering
        \includegraphics[width=\textwidth]{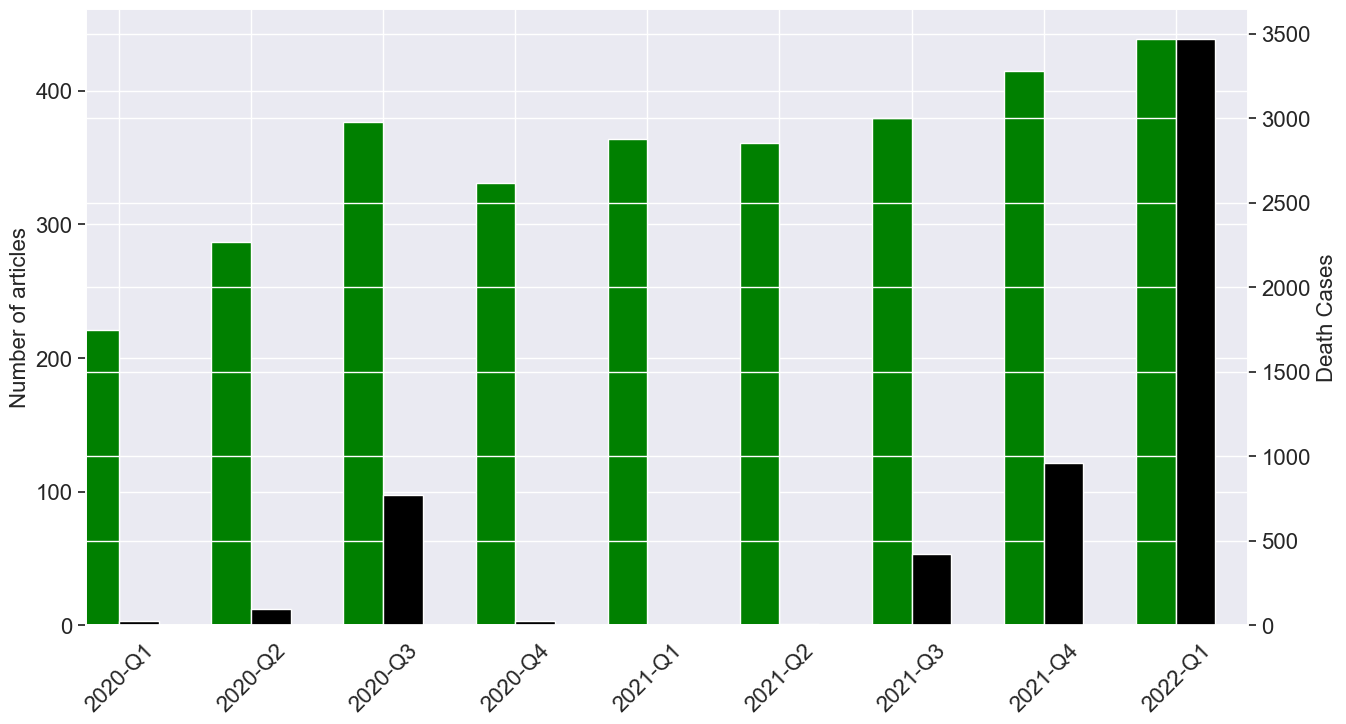}
        \caption{Australia data}
        \label{fig:ausdeathcases}
    \end{subfigure}
    \hfill
    \begin{subfigure}[b]{0.5\textwidth}
        \centering
        \includegraphics[width=\textwidth]{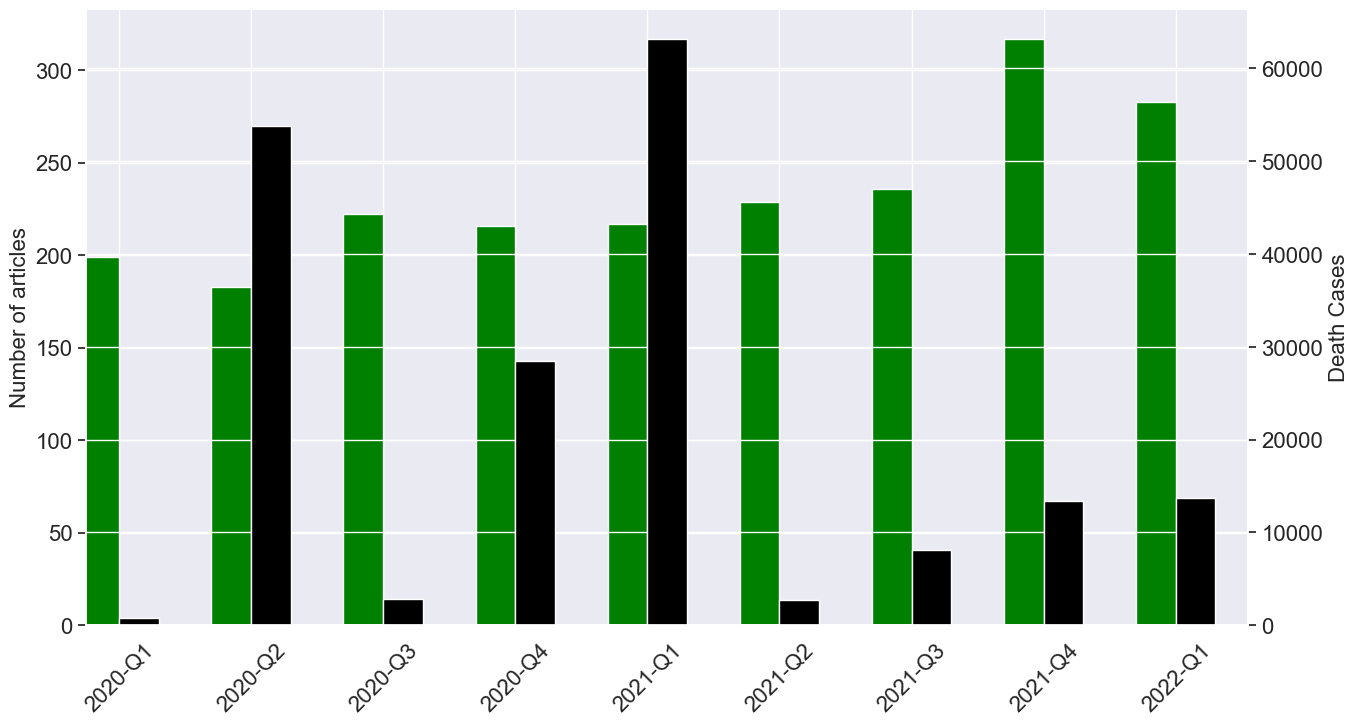}
        \caption{UK data}
        \label{fig:ukdeathcases}
    \end{subfigure}
    \caption{Comparative analysis of article content and death cases in  Australia and UK news sections of The Guardian.}
    \label{fig:compare_article_and_death_case}
\end{figure}

We first present statistics about COVID-19  deaths in relation to the  Australia and UK news sections of The Guardian, respectively.  Figure \ref{fig:compare_article_and_death_case} describes a comparative analysis of The Guardian article counts and number of death cases across nine quarters for the given years (from  the first quarter of 2020 to the first quarter of 2022). We extracted the COVID-19 death data  from the Australian Bureau of Statistics  \cite{a2023_covid19} and the UK's Office for National Statistics \cite{ons_2022_deaths}. In the case of Australia ( Figure \ref{fig:compare_article_and_death_case} - Panel (a)), we notice  the spikes in the COVID-19 death counts in the third quarter of 2020, the fourth quarter of 2021, and the first quarter of 2022 corresponds to an increase in the volume of news content, suggesting a correlation between the pandemic's severity and media coverage. In contrast, the UK experienced significant rises in death rates in the second quarter of 2020 and the first quarter of 2021, without an increase in the news report, indicating a possible divergence between public health data and media response. This analysis sets the stage for a focused examination of the specified quarters in subsequent, more detailed quarterly analyses.

\subsection{Data analysis using n-grams}

\begin{figure}[htbp!]  
    \centering
    \begin{subfigure}[b]{0.45\textwidth}
        \centering
        \includegraphics[width=\textwidth]{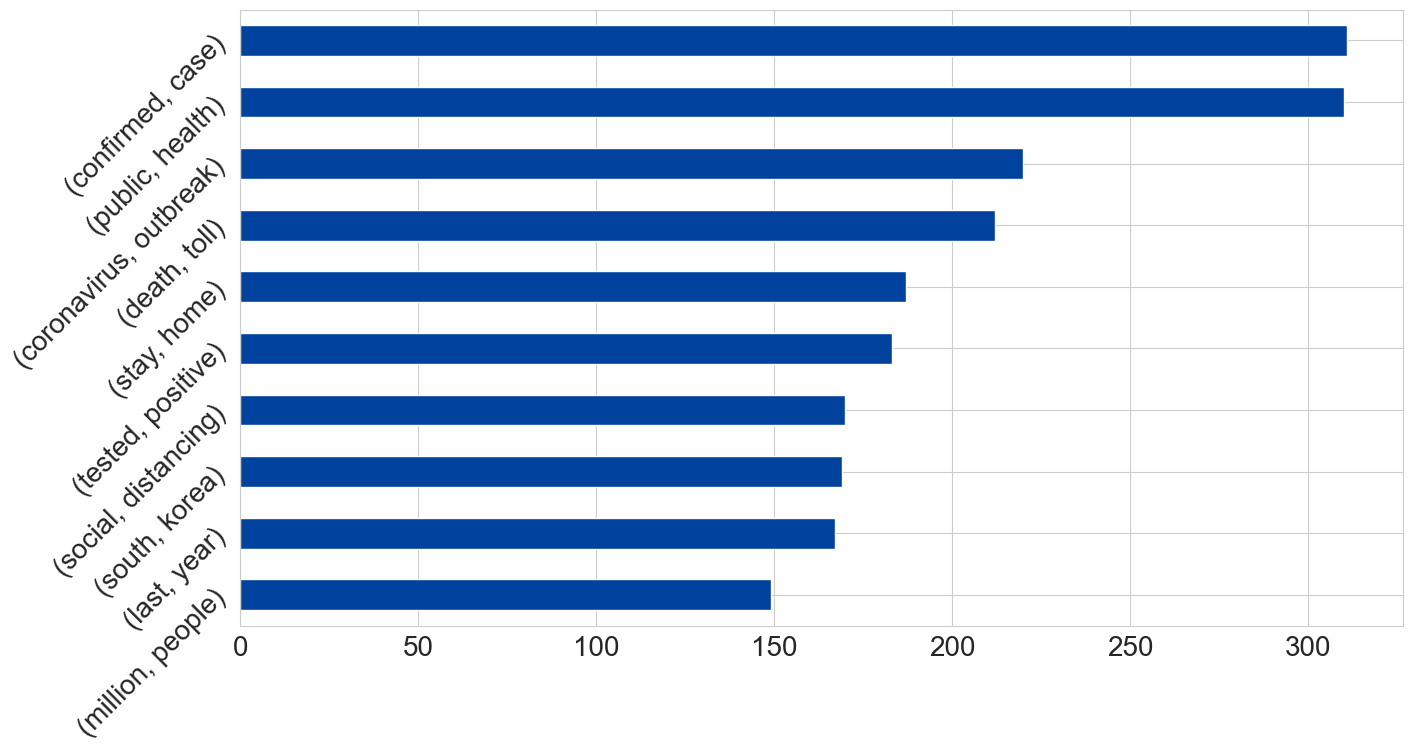}
        \caption{Bigrams}
        \label{fig:world_20q1}
    \end{subfigure}
    \hfill
    \begin{subfigure}[b]{0.45\textwidth}
        \centering
        \includegraphics[width=\textwidth]{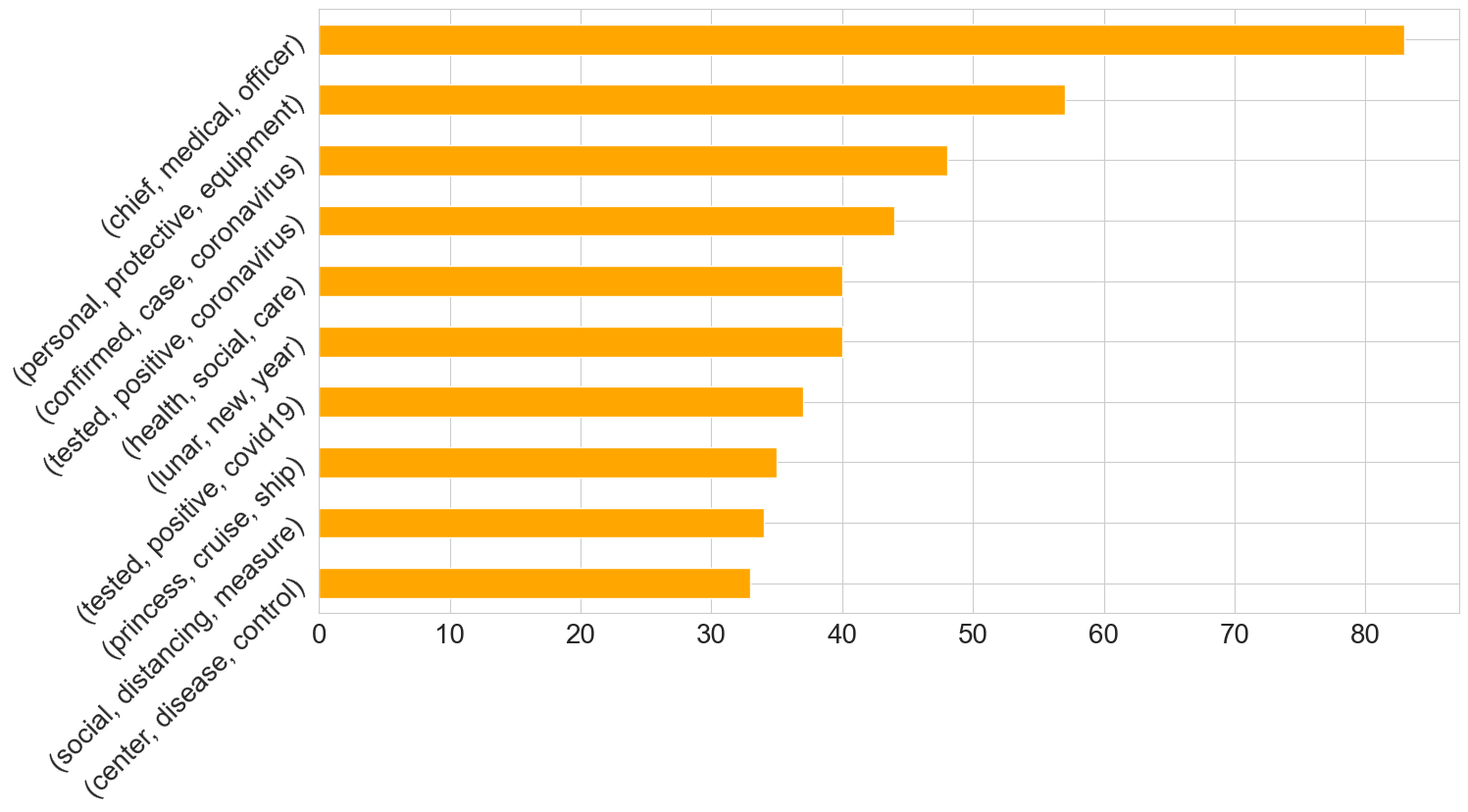}
        \caption{Trigrams}
        \label{fig:tri_world}
    \end{subfigure}
    \caption{Top 10 bigrams and trigrams for World News in 2020 first quarter (January  - March) covering the beginning of COVID-19 pandemic. 
    }
    \label{fig:bi_tri_world}
\end{figure}

%
Next, we use n-gram analysis to get an overview about the major topics expressed in the selected sections of The Guardian during the first phase of COVID-19.
According to Figure \ref{fig:bi_tri_world}-Panel (a), in the first quarter of 2020, the lexicon of global news was dominated by phrases deeply tied to the emerging COVID-19 pandemic, with "public health", "confirmed case", and "coronavirus outbreak" leading the frequency plots in bigrams analysis. The trigrams (Figure \ref{fig:bi_tri_world}-Panel (b)) show "chief medical officer", "public health England", and "confirmed case coronavirus" emerged as top phrases in the World News section. The phrase "confirmed case" in both bigrams and trigrams points to the critical importance of tracking the virus in the beginning of the pandemic. Similarly, the frequent appearance of "personal protective equipment" in discussions signals the pivotal role of protective gear in controlling the outbreak, hinting at the global surge in demand for such equipment and the supply chain challenges that ensued. This lexical analysis of early 2020 news coverage offers a window into the world's collective focus and concerns during the initial stages of the COVID-19 pandemic, revealing an acute awareness of the crisis's impact on public health.


\begin{figure}[htbp!]

    \centering
    \begin{subfigure}[b]{0.45\textwidth}
        \centering
        \includegraphics[width=\textwidth]{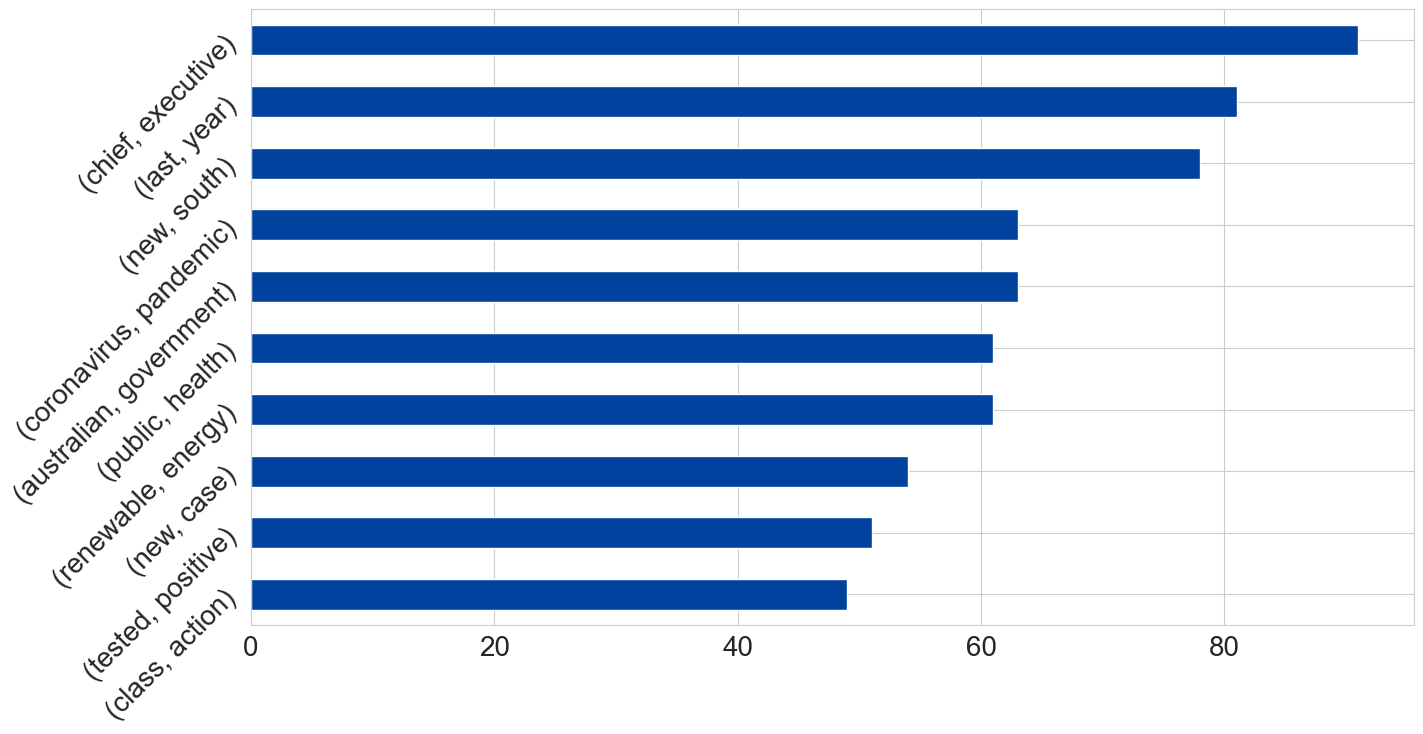}
        \caption{Australia news}
        \label{fig:au20q2bi}
    \end{subfigure}
    \hfill
    \begin{subfigure}[b]{0.45\textwidth}
        \centering
        \includegraphics[width=\textwidth]{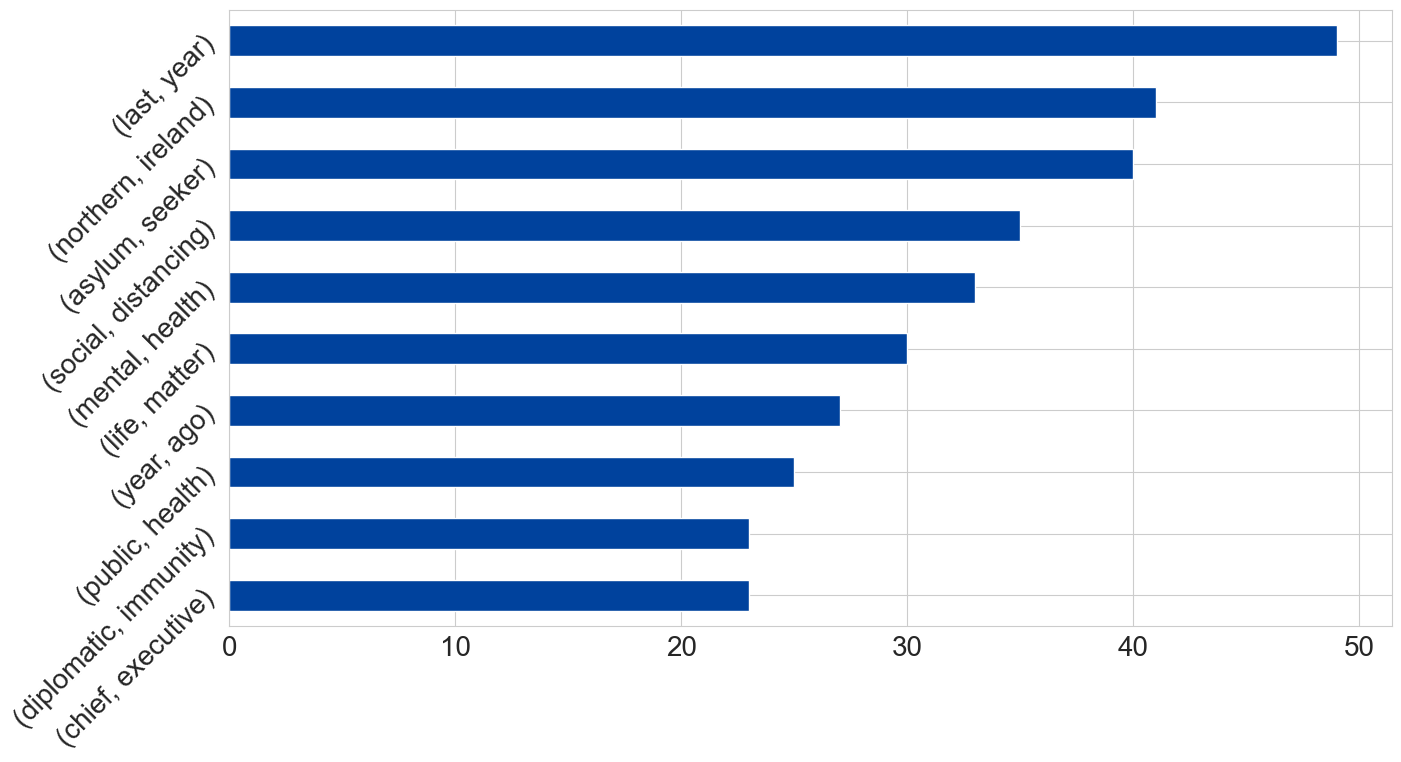}
        \caption{UK news}
        \label{fig:uk20q2bi}
    \end{subfigure}
    \caption{Top 10 bigrams for Australia news and UK news in 2020 second quarter (April - June) covering the beginning of lockdowns of the COVID-19 pandemic
    }
    \label{fig:bi_au_uk_q2}
\end{figure}

In the second quarter of 2020 (Figure \ref{fig:bi_au_uk_q2}), analysis of bigrams in Australian and UK news highlights key concerns related to the COVID-19 pandemic. We find that Australia focused on "coronavirus pandemic", "public health", "new case", and "tested positive", reflecting a close attention to the outbreak's evolution and its health implications. In contrast, the UK emphasised "social distancing", "mental health", and "public health", pointing to a broader scope of concerns that included preventive measures and the psychological impact of the pandemic. This contrast reveals differing priorities, where Australia's media concentrated on tracking and testing the virus, whereas the UK highlighted the social and mental health dimensions of the crisis. Through this lens, we can glimpse how each country's media spotlighted shared and distinct aspects of the pandemic, underscoring diverse strategies and responses to the pandemic.

\subsection{BERT and RoBERTa - model evaluation}

We present the training (model refinement) performance metrics for both the BERT and RoBERTa models using the SenWave training dataset with 90:10 percent train and test data split. The  given metrics encompass  Hamming loss, Jaccard score, LRAP score, macro and micro F1 scores. These key performance metrics are given in the literature \cite{tsoumakas2007multi} and used in related work for multi-label sentiment classification \cite{chandra2021covid}. They offer a comprehensive evaluation of the model's effectiveness in multi-label sentiment classification. Hamming loss measures incorrect label predictions, where lower values signify better performance. The Jaccard score assesses the similarity between predicted and true labels, with higher values indicating better overlap. The LRAP score evaluates the ranking quality of predictions based on the average precision of true labels, with higher values indicating superior ranking performance. The F1 macro-score computes the average F1 score across all classes, reflecting precision and recall equally. Similarly, the F1 micro-score considers the overall F1 score, useful for addressing class imbalance. The higher values of the respective F1 scores signify better model performance. 

 We refine the respective models using the Senwave train dataset and report the performance on the test dataset. We present the results in  Table \ref{tab:metrics}, where RoBERTa exhibits marginally superior performance across various metrics, excelling in LRAP, Jaccard and F1 micro scores. This indicates a slightly better overall performance when compared to the BERT model.

\begin{table}[H]
\centering
\begin{tabular}{|c|c|c|}
\hline
\textbf{Metric} & \textbf{BERT} & \textbf{RoBERTa} \\ 
\hline
Hamming loss & 0.147 & 0.135 \\
\hline
Jaccard score & 0.496 & 0.519 \\
\hline
LRAP score & 0.753 & 0.774 \\
\hline
F1 Macro-score & 0.536 & 0.533 \\
\hline
F1 Micro-score & 0.576 & 0.591 \\
\hline
\end{tabular}
\caption{Refining BERT and RoBERTa models using SenWave test dataset.}
\label{tab:metrics}
\end{table}


\subsection{Sentiments detected by LLMs}

\begin{figure}[htbp!]
    \centering
    \includegraphics[width=0.5\textwidth]{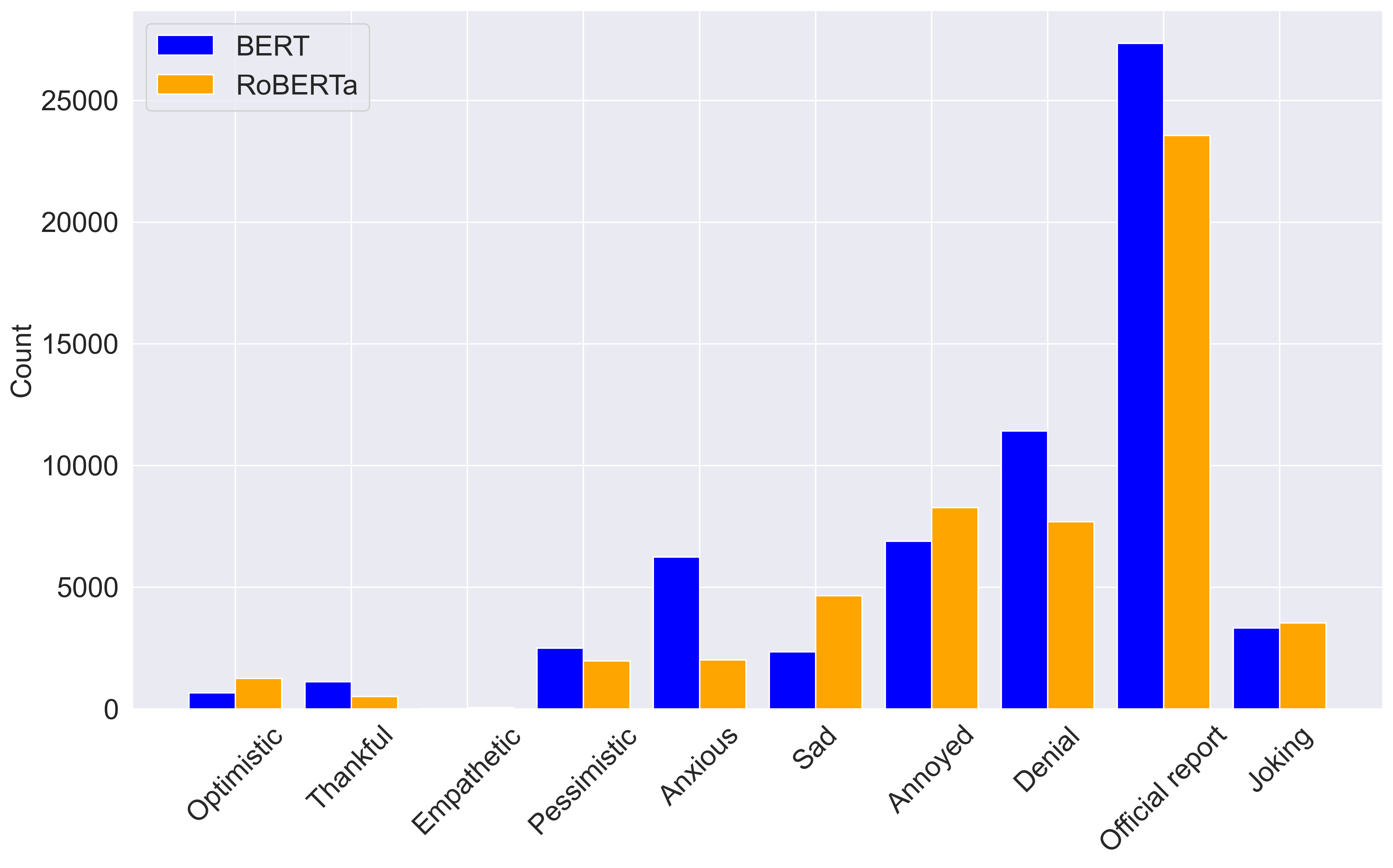}
    \caption{Sentiment distribution  by the BERT and RoBERTa models from The Guardian (Australia, UK, World news and Opinion sections combined), spanning from 1st January, 2018, to 31st March, 2022.}
    \label{fig:senBERTvsRoBERTa}
\end{figure}

We next compare the sentiments detected by BERT and RoBERTa models. 
Figure \ref{fig:senBERTvsRoBERTa} presents sentiment distribution within The Guardian (Australia, UK, World news and Opinion sections), spanning from January 1, 2018, to March 31, 2022. This provides the predominant sentiment categories as identified by BERT and RoBERTa  offering insights into the overarching emotional tone of the media coverage.
A significant finding from Figure  \ref{fig:senBERTvsRoBERTa} is the prevalence of "official report" as the dominant sentiment across the dataset, regardless of the LLMs used. This observation aligns with the nature of news reporting, which is fundamentally centred on neutrality and objectivity. Certain news articles  conveyed information from official statements and reports during COVID-19. Therefore, the prominence of "official report" as a sentiment category underscores the adherence of the news media to these journalistic standards. Although  the prevalence of "official report" sentiment is expected given the context of news reporting, it presents a challenge for our analysis focused on capturing a broader spectrum of sentiments related to the COVID-19 pandemic. The overrepresentation of this category may potentially obscure the nuances of public sentiment and emotional responses to the pandemic as reflected in news coverage. Consequently, to address this issue and enhance the clarity of human sentiment expressed, we excluded "official Report" from subsequent analyses.
This will allow for a more nuanced exploration of the emotional landscape of news coverage during the specified period. We further aim to uncover more distinct emotional patterns and shifts that may provide deeper insights into the collective sentiment towards the pandemic, beyond the confines of official reporting. Finally, we notice that  the "official report", "annoyed," and "denial" make the major differences between the sentiment counts when comparing BERT with RoBERTa in Figure \ref{fig:senBERTvsRoBERTa}.

\begin{figure}[htbp!]
    \centering
    \includegraphics[width=\linewidth]{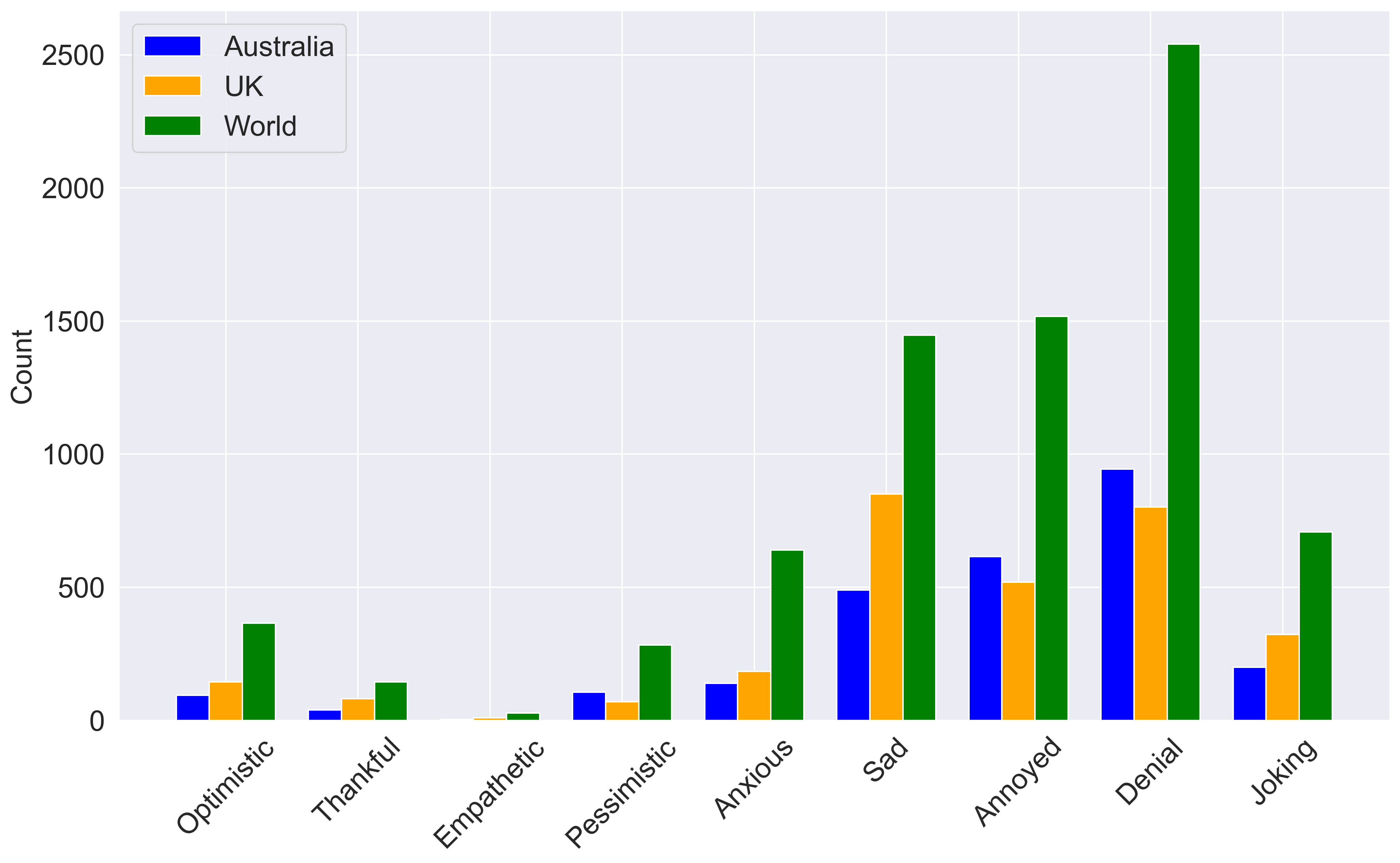} 
    \caption{Sentiments detected in Australia, UK and World news sections cover 1st January, 2018 - 31st March, 2022.}  

    \label{fig:em_dis_aus_uk_world}
\end{figure}

Now that we compared the BERT and RoBERTa models, we will present the rest of the analyses using RoBERTa as our designated LLM. 

Figure \ref{fig:em_dis_aus_uk_world} displays the distribution of emotions within Australian, UK, and World news sections, providing an intriguing perspective on how location-specific contexts influence sentiment portrayal. Despite the smaller dataset for the UK, with a thousand fewer articles than in Australia, an interesting pattern emerges. The sentiment "sad" is more pronounced in UK coverage, which could potentially reflect the higher COVID-19 mortality rates experienced in the UK compared to Australia during the pandemic since the data also covers the COVID-19 period. Moreover, it's noteworthy that sentiments such as "optimism", "thankful", and even "joking" are more prevalent in the UK news compared to Australia. Both UK and Australia have reported far less  negative sentiments (sad, annoyed, anxious, and denial) when compared to the World News section. This phenomenon might indicate a cultural or editorial leaning towards maintaining a semblance of hope and gratitude, even when faced with dire circumstances. The higher occurrence of these positive emotions, despite the adversity, aligns with our previous observations of an increase in such sentiments during the pandemic period. Furthermore, it could also indicate a political bias of The Guardian in reporting negative news for the rest of the world, and more positive news about the Western world (UK and Australia).

\begin{figure}[htpb!]
    \centering
    \includegraphics[width=0.45\textwidth]{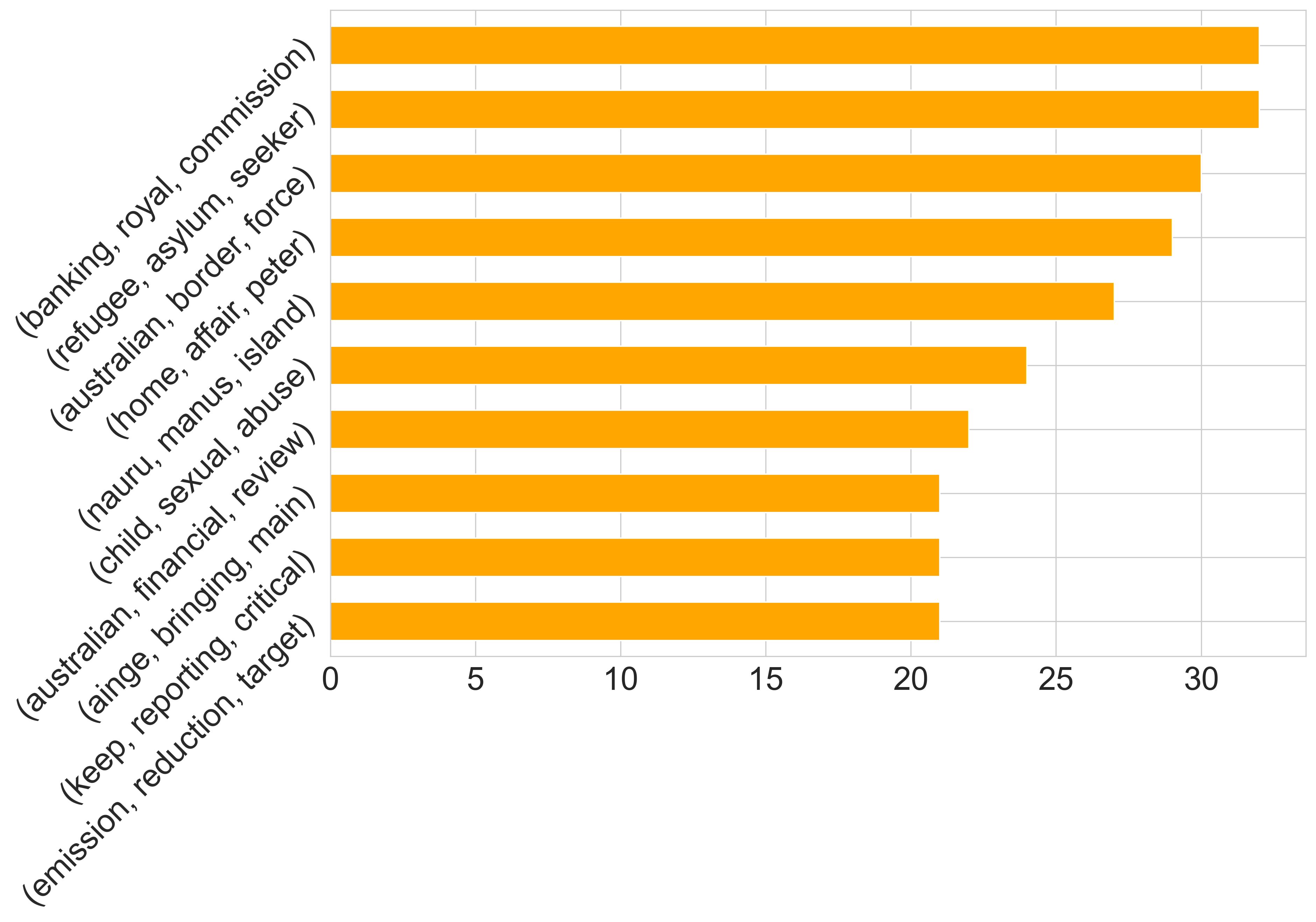}
    \caption{Top 10 trigrams of the Australia news before COVID-19.}
    \label{fig:aupre}
\end{figure}

 In Figure \ref{fig:aupre}, we filtered articles from the Australia news section spanning January 1, 2018, to December 31, 2019, each of which contained sentiments such as "sad", "annoyed", and "denial". We selected this based on the three emotions with the highest proportion in 2018 and 2019 in Figure \ref{fig:quarterly_emotion_disausroberta}. Through this analysis, we discovered that these trigrams encompass various topics, including economics, politics, and military affairs.

\subsection{Sentiment polarity analysis}

In the global impact of COVID-19, media reporting on pandemic-related information was influenced by various factors, including the  government response measures, and public opinion. Consequently, media coverage would exhibit varying sentiments depending on the nature and subject of the news coverage. Analysing changes in sentiment polarity in Guardian articles before and during COVID-19 can provide an insight of the sentiments in the different phases. 

We assigned weights to each sentiment to compute our custom sentiment polarity scores for the Guardian articles using RoBERTa, as presented in Table \ref{tab:weights}. These scores range from -1 to 1, where a positive score denotes favourable sentiment towards the statement, and a negative score suggests adverse sentiment.

\begin{table}[htbp]
\centering
\begin{tabular}{|l|c|}
\hline
\textbf{Sentiment} & \textbf{Weight} \\
\hline
Optimistic & 3 \\
Thankful & 2 \\
Empathetic & 0 \\
Pessimistic & -3 \\
Anxious & -2 \\
Sad & -2 \\
Annoyed & -1 \\
Denial & -4 \\
Official report & 0 \\
Joking & 1 \\
\hline
\end{tabular}
\caption{Sentiments and weights.}
\label{tab:weights}
\end{table}

\begin{figure}[H]
    \centering
    \includegraphics[width=0.5\textwidth]{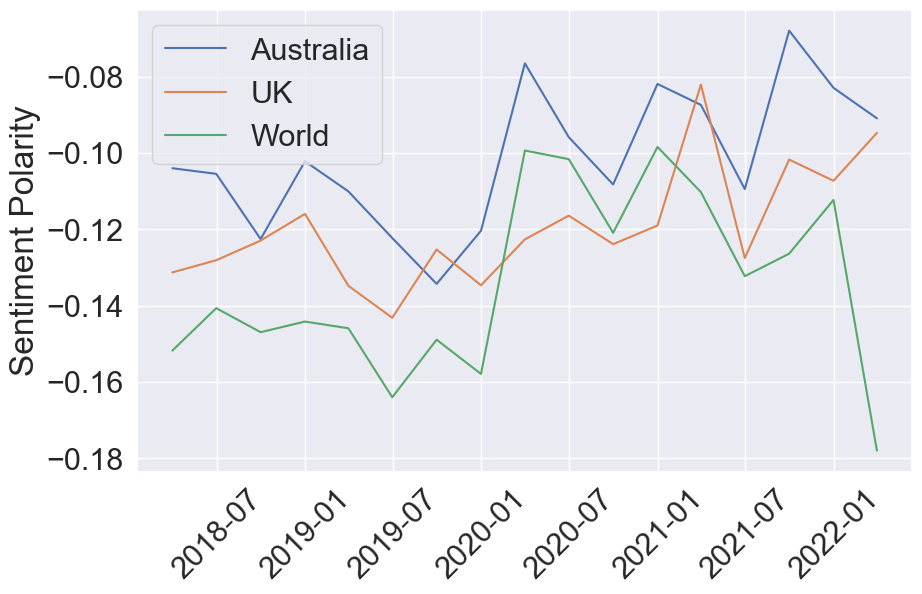}
    \caption{Selected regions' quarterly polarity score fluctuation.}
    \label{fig:sentimentpolarity}
\end{figure}
Sentiment polarity score in these sections of The Guardian increased significantly from 2019 to 2020, at the start of COVID-19, as shown in Figure \ref{fig:sentimentpolarity}. According to The Guardian (2020) \cite{a2020_the}, this surge can be attributed to increased empathy, positive media narratives and people’s ability to adapt. People may write about solidarity, hope, and inspiration despite challenges, creating an upward trend of overall mood improvement.

\begin{figure}[htbp!]
    \centering
    \includegraphics[width=0.5\textwidth]{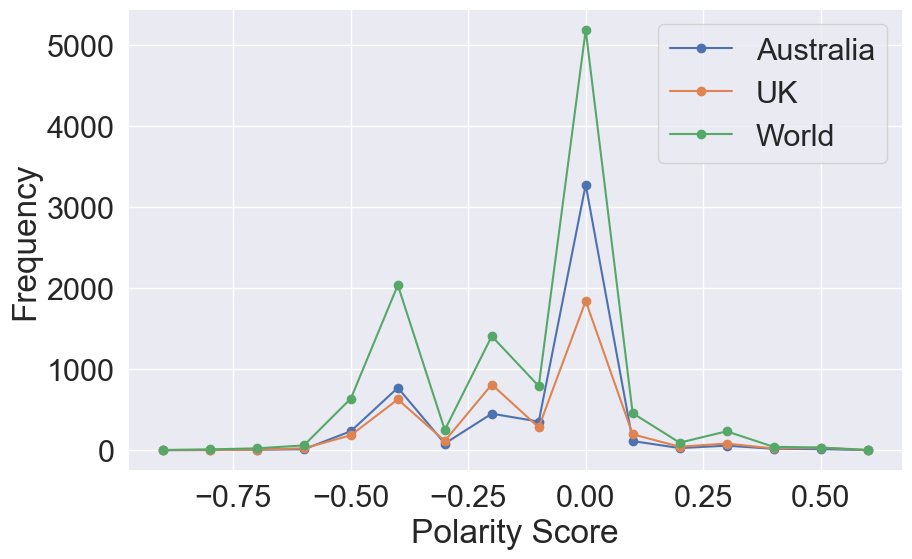}
    \caption{Distribution of polarity scores for different regions for COVID-19 pandemic  from 1st Jan 2018 to 31st March 2022}.
    \label{fig:distributionsp}
\end{figure}

According to Figure \ref{fig:sentimentpolarity}, in the overall analysis, the polarity of pre-COVID-19 and post-COVID-19 articles tends to lean towards negative sentiment, which may be attributed to the higher proportion of negative sentiment in the weights we assign. We focus on data from the first quarter (January - March) 2018 to the first quarter of 2022,  including Australia and the UK sections. During this period, articles in all three sections of The Guardian showed swings in emotional polarity, reflecting the emotional ups and downs experienced by individuals during the COVID-19 pandemic. As shown in Figure \ref{fig:distributionsp}, the sentiment polarity of The Guardian is heavily skewed toward the negative, indicating the prevalence of emotional distress among individuals in the COVID-19 era.

Looking at the Australian section of  The Guardian in Figure \ref{fig:sentimentpolarity}, the polarity showed a downward trend from the second quarter of 2020 to the fourth quarter of 2020, possibly due to the rapid increase in global epidemic cases during this period. Sentiment polarity then continued to swing downward towards the last quarter of 2021. This quarterly decline may be related to Australia's surge in new cases over the same period and current socioeconomic conditions. Instead, the rebound in sentiment polarity in the UK in the first quarter of 2022 may be attributed to the stabilisation of the pandemic in the country  as shown in Figure \ref{fig:compare_article_and_death_case}. The overall trend in the global sentiment polarity fluctuated downward throughout the COVID-19 period, particularly evident in the first quarter of 2022, suggesting a potential negative impact of the epidemic on global emotional well-being. The regional differences in sentiment polarity values may be associated with factors such as epidemic response measures, change in vaccination rates, and economic recovery efforts in each region.

\begin{figure}[htbp]
    \centering 
    \begin{subfigure}{0.45\textwidth}
\includegraphics[width=\linewidth]{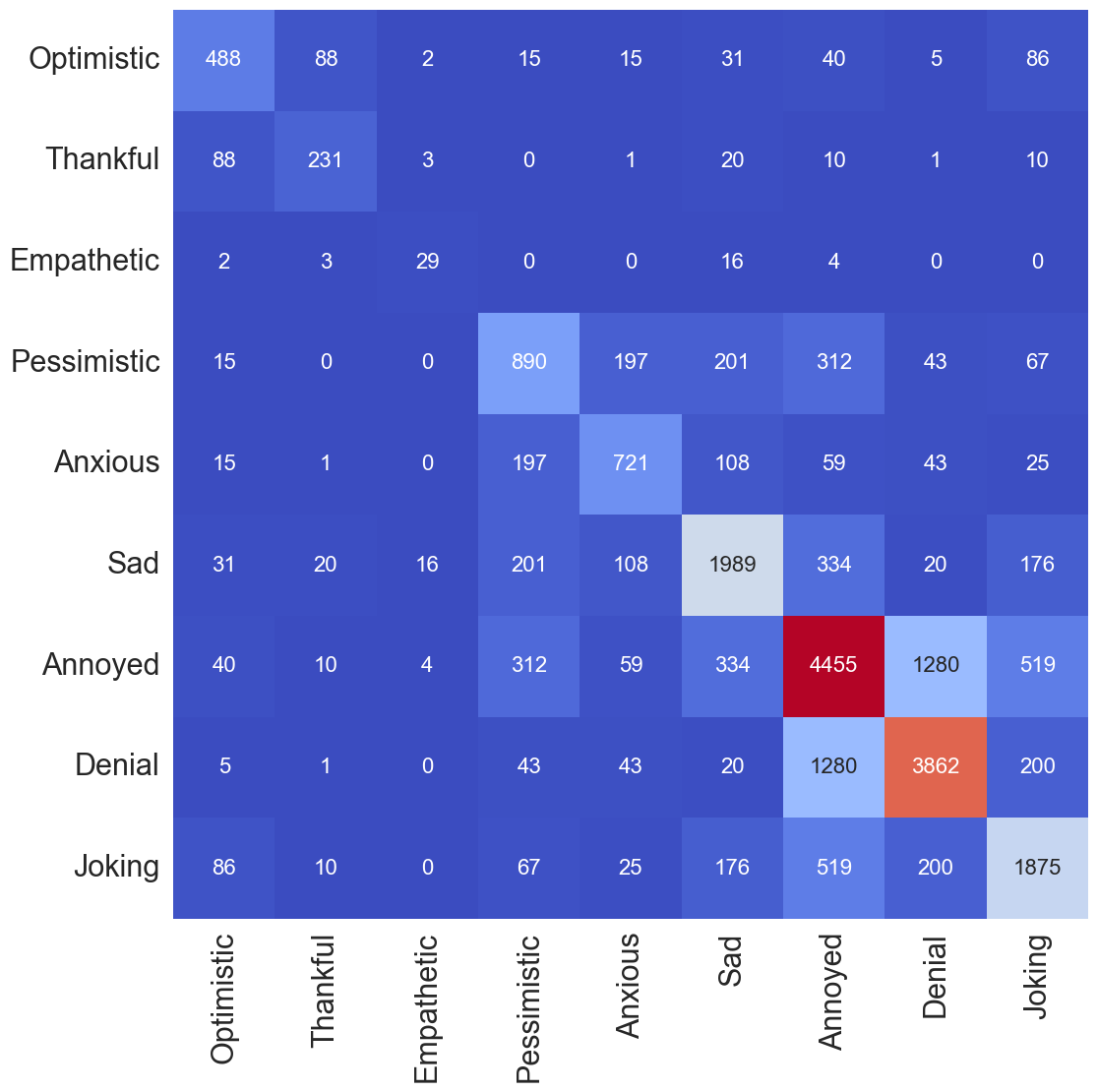}
        \caption{Pre-COVID-19}
        \label{fig:heatprer}
    \end{subfigure}
    \hfill
    \begin{subfigure}{0.45\textwidth}
\includegraphics[width=\linewidth]{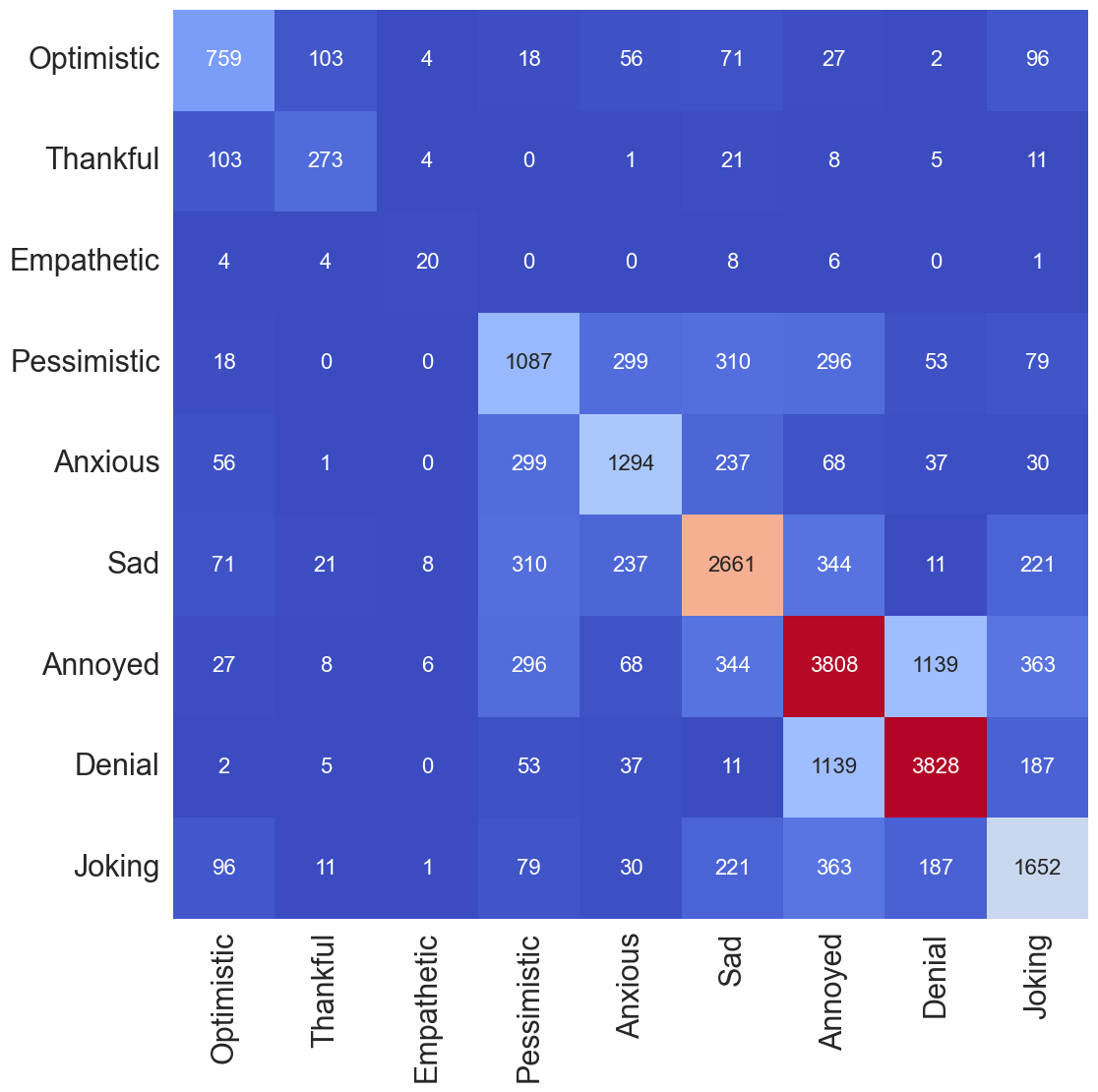}
        \caption{During COVID-19}
        \label{fig:heatdurr}
    \end{subfigure}
    
    \caption{Heatmap showing the relationship between different sentiments (a) before and (b) during COVID-19 using RoBERTa model.  Note the data features Australia, UK, World News and Opinion sections.}
    \label{fig:BERTvsRoBERTa-heatmap}
\end{figure}

\begin{figure}[htbp!]
    \centering
    \includegraphics[width=0.5\textwidth]{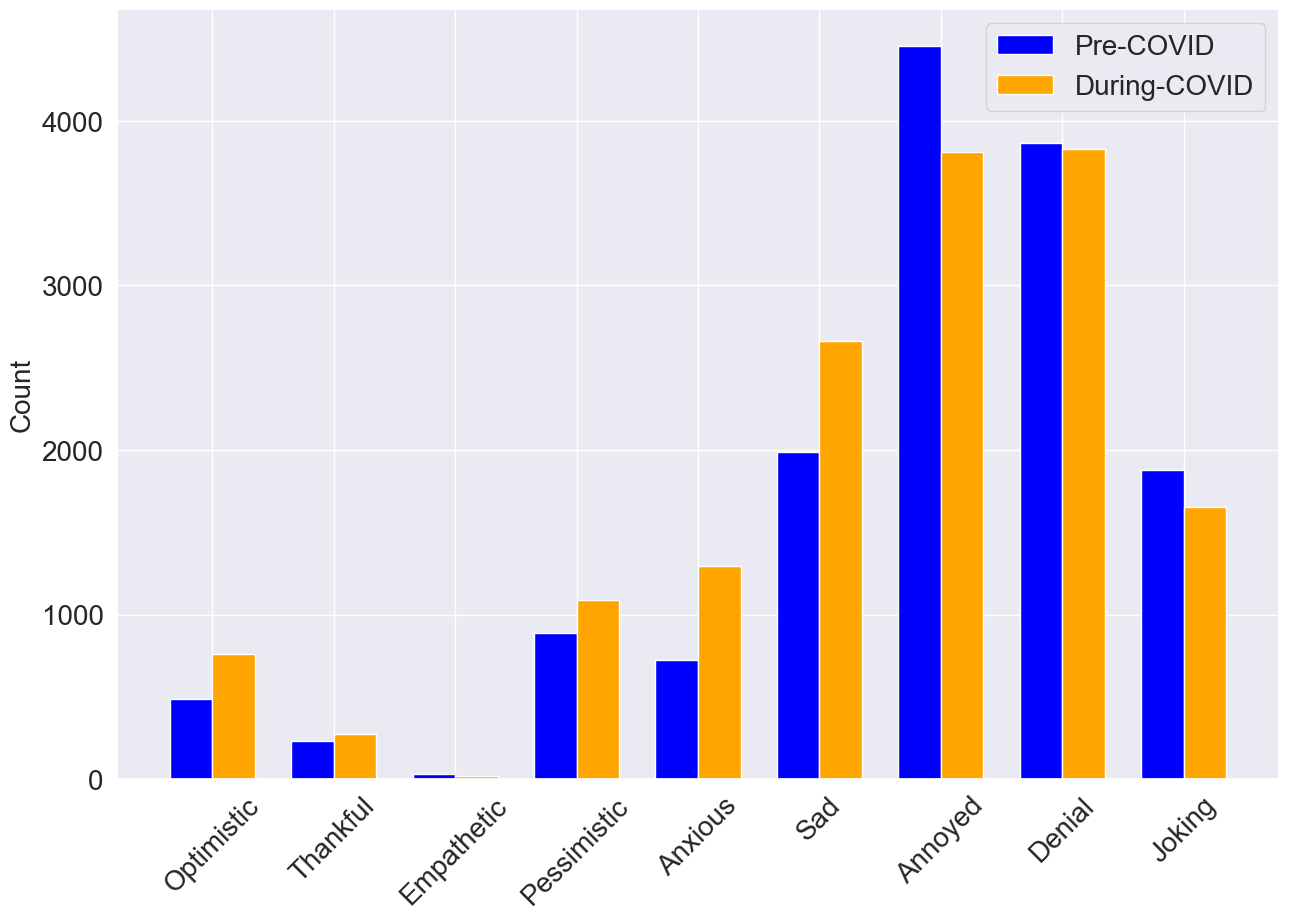} 
    \caption{Sentiments detected before and during COVID-19 using RoBERTa model with data fromAustralia, UK, World News, and Opinion sections combined.}
    \label{fig:Pre-COVIDvDuring-COVID_roberta}
\end{figure}

\subsection{Sentiments differences pre-COVID-19 and during COVID-19}

We need to have an understanding of the impact of COVID-19 on the style of the news coverage by The Guardian and this can be done by comparing the sentiment analysis results pre-COVID-19 and during COVID-19.
Figures \ref{fig:BERTvsRoBERTa-heatmap} and \ref{fig:Pre-COVIDvDuring-COVID_roberta} present sentiment distribution and correlation using the RoBERTa model, spanning across the pre-pandemic and pandemic periods. It is evident from the analysis that both periods are predominantly characterized by negative sentiments. Specifically, "denial", "annoyed", and "anxious" were the most frequently occurring sentiments detected by the BERT model.  Conversely, RoBERTa identified "annoyed" as the most common sentiment, followed closely by "denial" and "sad". The onset of the pandemic brought about a noticeable increase in "anxious" and "sad" across both models, reflecting the global uncertainty and grief triggered by the crisis. This shift underscores the deep emotional impact the pandemic has had on societal sentiments, resonating through the tone of news coverage. Interestingly, despite the surge in negative sentiments, the models also registered an uptick in positive sentiments such as "optimism" and "thankful" during the pandemic. 

\begin{figure}[htbp!]
    \centering
    \begin{subfigure}[b]{0.45\textwidth}
        \centering
        \includegraphics[width=\textwidth]{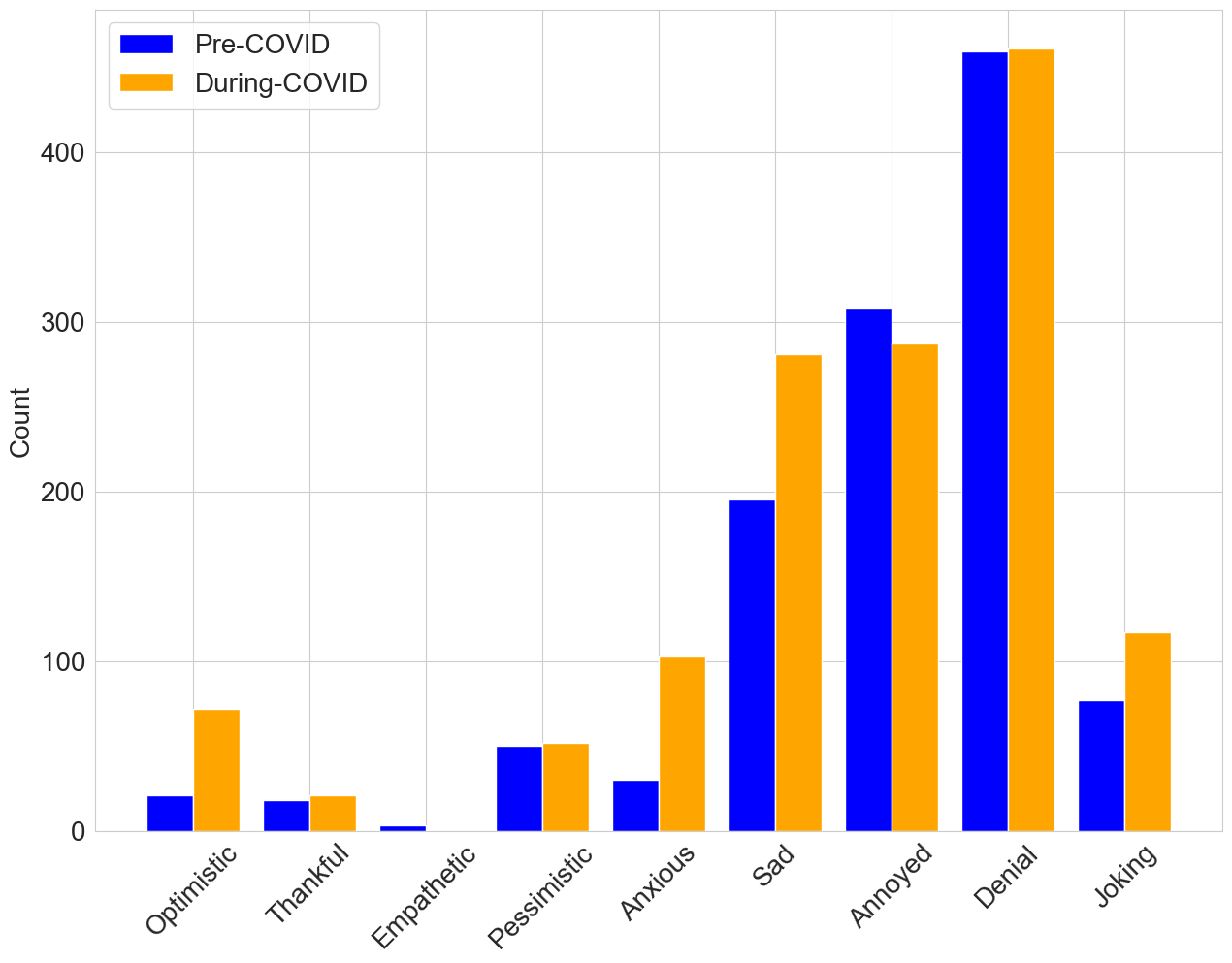}
        \caption{ Australia news}
        \label{fig:aupd}
    \end{subfigure}
    \hfill
    \begin{subfigure}[b]{0.45\textwidth}
        \centering
        \includegraphics[width=\textwidth]{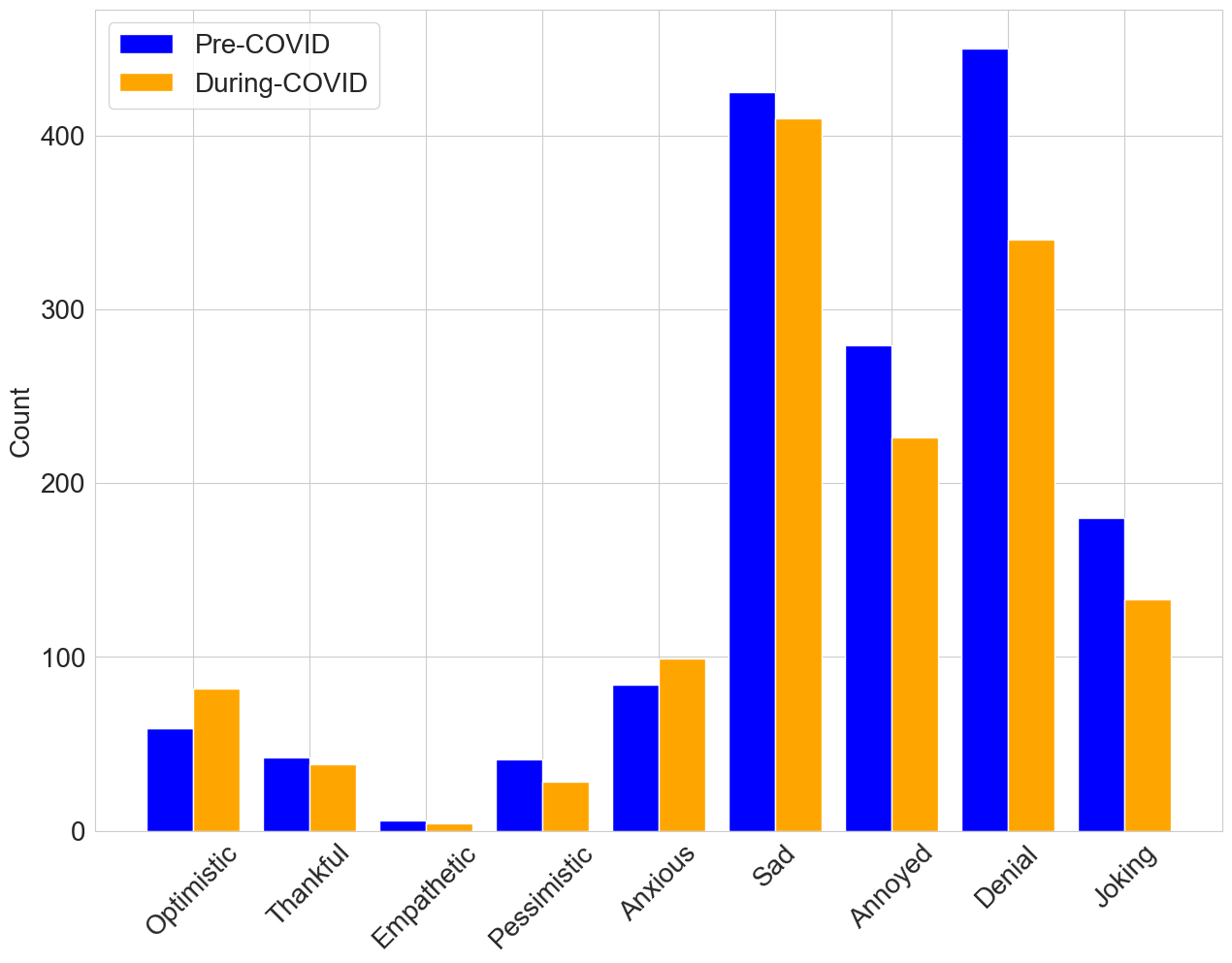} 
        \caption{ UK news}
        \label{fig:ukpd}
    \end{subfigure} 
    \hfill
    \begin{subfigure}[b]{0.45\textwidth}
        \centering
        \includegraphics[width=\textwidth]{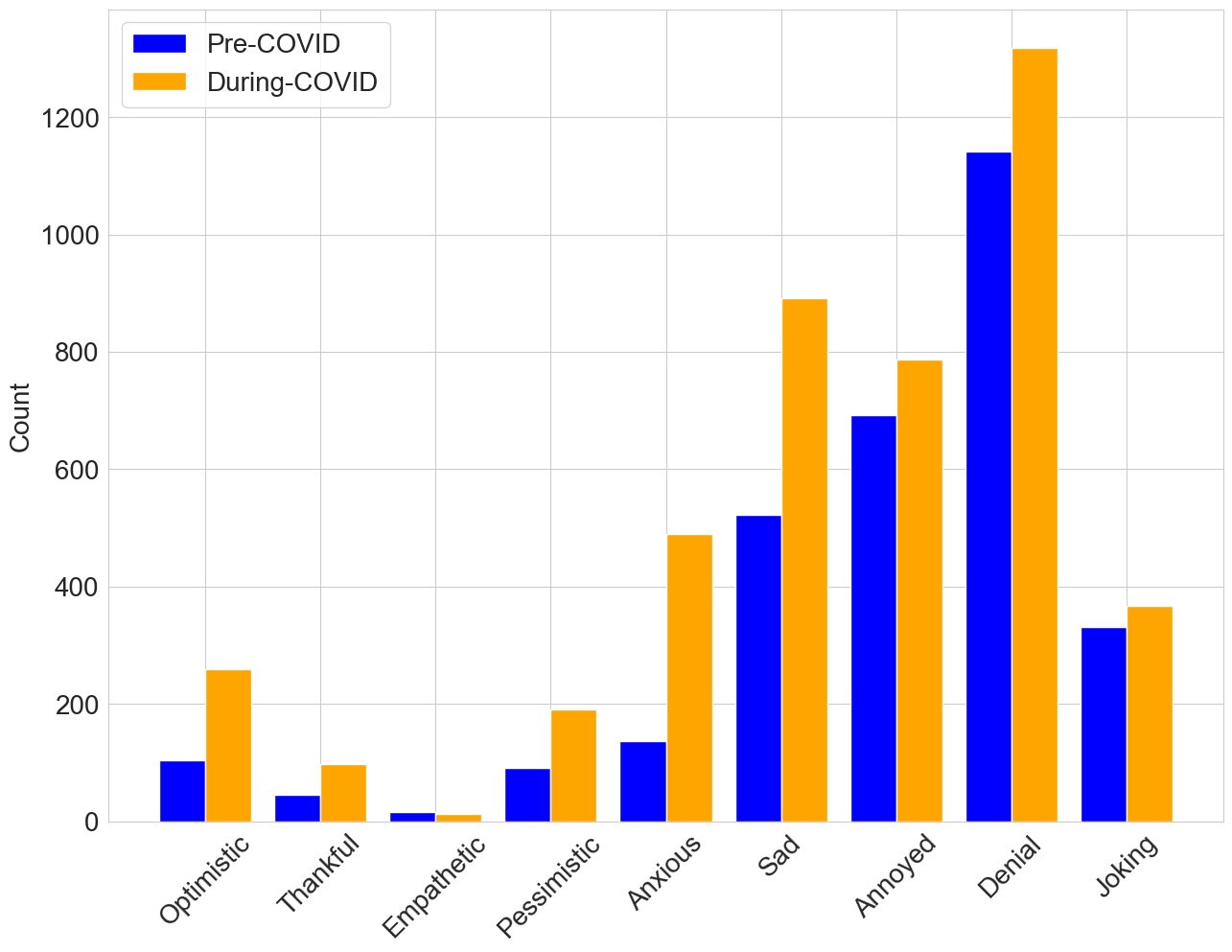} 
        \caption{World news}
        \label{fig:wopd}
    \end{subfigure}
    \caption{Sentiments detected in  the Australia, UK and World News sections, before and during COVID-19.}
    \label{fig:before_after_auw}
\end{figure}

In Figure \ref{fig:before_after_auw}, we can easily see that negative sentiments such as "denial", "sad" and "annoyed" are the most prominent in Australia, UK and World News sections, before and during COVID-19. Comparing the sentimental changes before and during COVID-19, the content of negative sentiments in Australia and the World News has increased compared to before the pandemic. On the contrary, in UK news, the proportion of negative sentiments during the pandemic has decreased.

\subsection{The Guardian's Opinion section}
\begin{figure}[htbp!]
    \centering
    \includegraphics[width=0.5\textwidth]{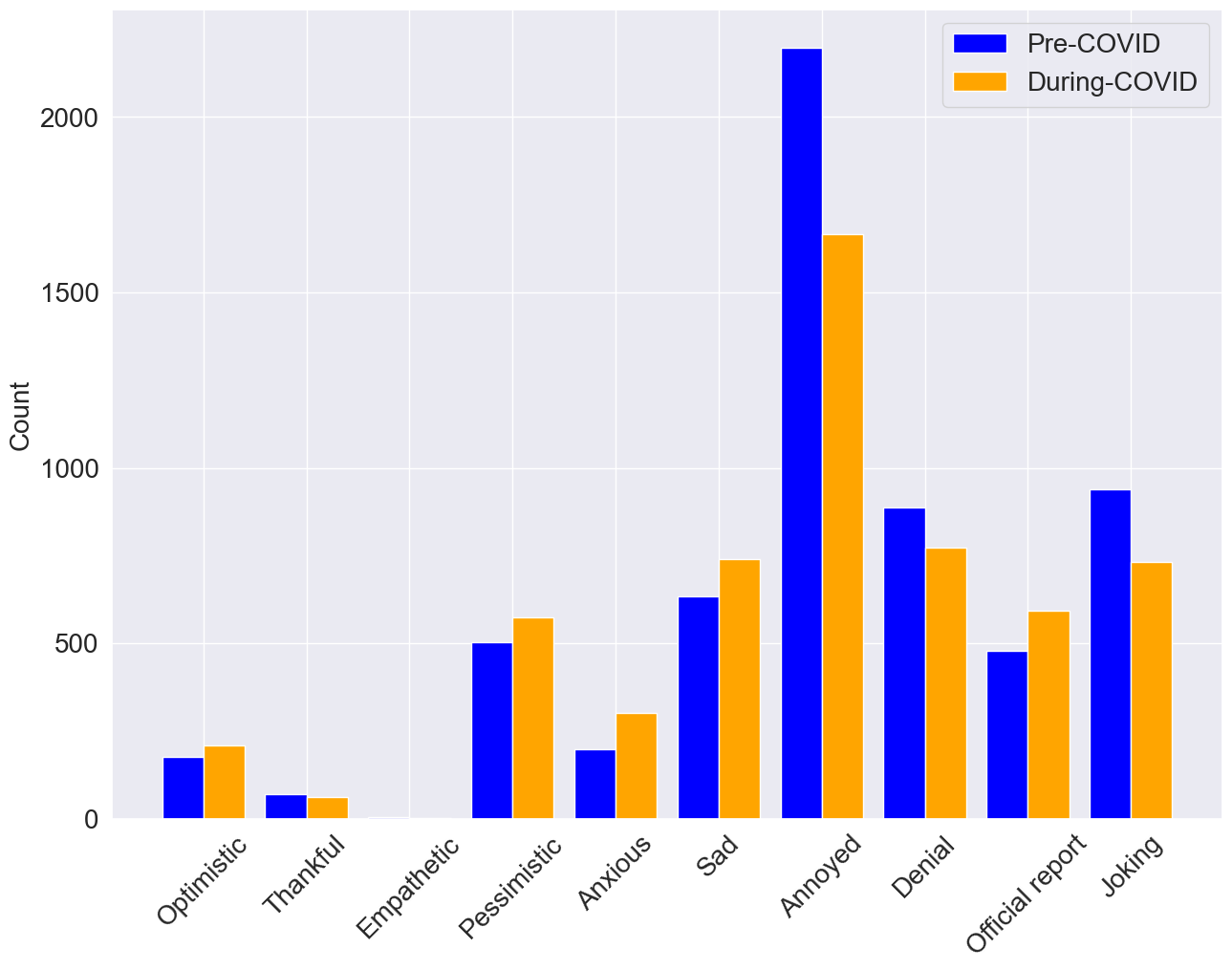}
     
    \caption{Sentiments detected in the  Opinion section before and during COVID-19.}  
    \label{fig:em_dis_opinion}
\end{figure}

We transition to a focused examination of the sentiments from  the Opinion section of The Guardian presented in Figure \ref{fig:em_dis_opinion}. We reintegrate the "official report" sentiment which was previously omitted from earlier analyses to focus on actual sentiments. As mentioned earlier, the "official report" category typically dominates due to the expected neutrality in news reporting. However, the Opinion section deviates from this trend, with "annoyed" and "denial" being the leading sentiments instead of "official report". This distinctive distribution aligns with the nature of opinion pieces, which often serve as a platform for expressing dissent and subjective viewpoints.The analysis of the Opinion section across the pre-pandemic and during-pandemic periods reveals surprising shifts in sentiment. Contrary to what might be expected, "annoyed" and "denial" sentiments exhibit a decrease during the pandemic, while "anxious" and "sad" sentiments show a justifiable increase, reflecting the global distress caused by the crisis. Concurrently, the increase in "official report" sentiments during the pandemic is also understandable, as opinion writers often refer to official statements and statistics to ground their arguments amidst the unfolding events. Interestingly, the "joking" sentiment has decreased suggest a shift towards more serious discourse in opinion pieces during the pandemic, indicating the gravity of the situation and possibly a change in the role of humour in public commentary. Although these shifts are noticeable, they are not marked by a significant magnitude. This could indicate that while the emotional tone of opinion pieces has shifted to accommodate the realities of the pandemic, the fundamental nature of the opinion section as a space for personal expression remains intact. It may also suggest that opinion writers and readers have maintained a level of emotional consistency, despite the transformative global events.
 


\subsection{Sentiment-based trigram analysis}

\begin{figure}[htpb!]
    \centering
    \includegraphics[width=0.45\textwidth]{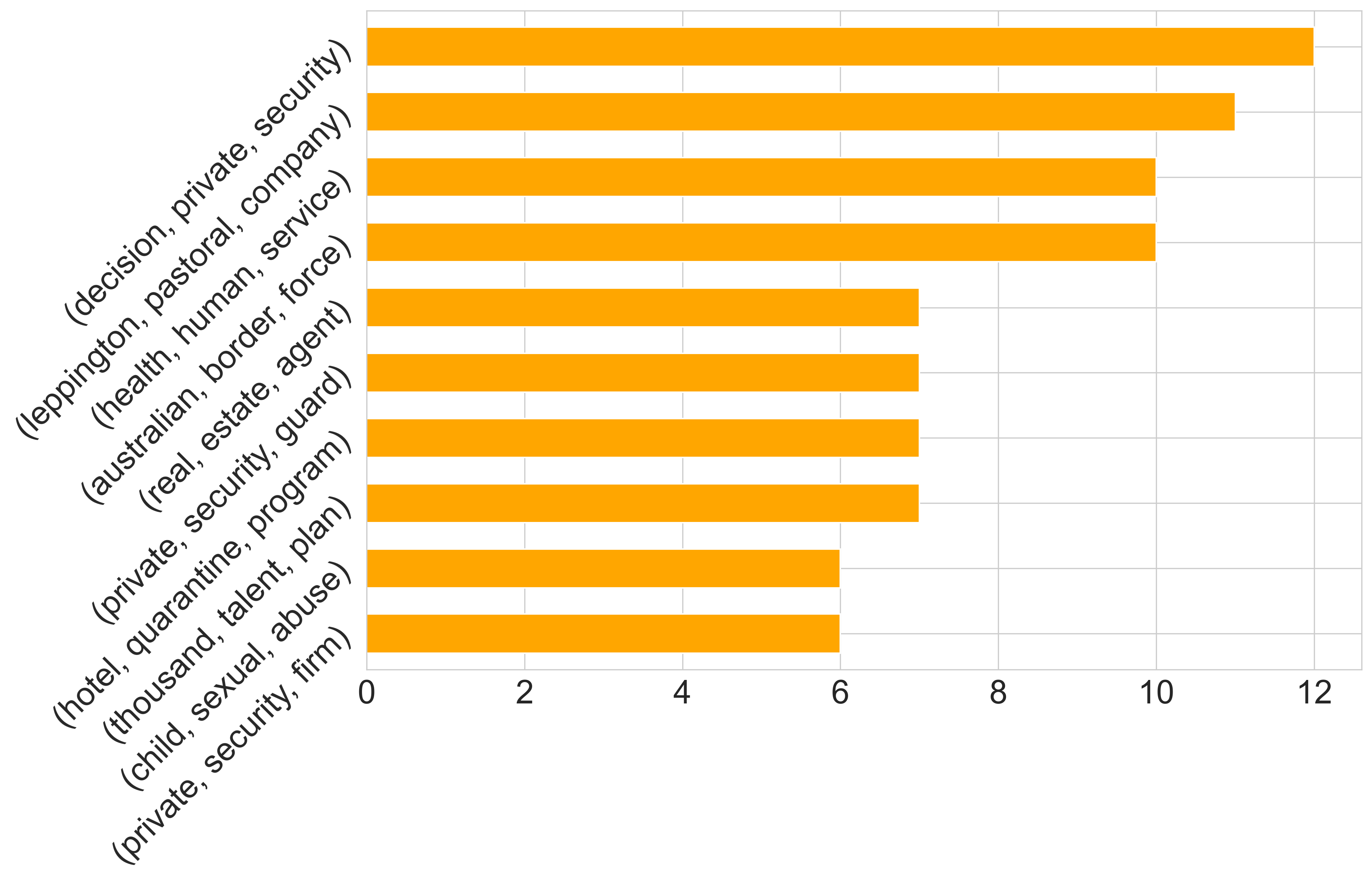}
    \caption{Top 10 trigrams of the Australia news during 2020 third quarter (July - September) for sentiments "sad, annoyed and denial".}
    \label{fig:au20q3}
\end{figure}

We have seen the trend pre and during COVID-19 and different sentiments expressed, but this does not give an indication of why such sentiments were expressed. A way to know why the sentiments such as "denial" or "annoyed" were expressed would be by looking at the texts where the LLM detected such sentiments and providing trigram analysis. Hence, this is a way backward, where we look at sentiments detected and then analyse the text further for those sentiments. In Figure \ref{fig:ausdeathcases}, a minor peak in the number of death cases in Australia during the early stages of COVID-19 is evident, occurring in the third quarter of 2020. Guided by the prevalent emotions identified in Figure \ref{fig:Aq3r} during the same period, which notably included "sad", "annoyed", and "denial", we selected the Australia news section and extracted trigrams. Figure \ref{fig:au20q3} presents phrases such as "health human service" and "hotel quarantine program" within these trigrams. "Health human service" likely pertains to medical services during the COVID-19 crisis, while "hotel quarantine program" likely relates to isolation policies implemented during the epidemic.

\begin{figure}[htpb!]
    \centering
    \includegraphics[width=0.45\textwidth]{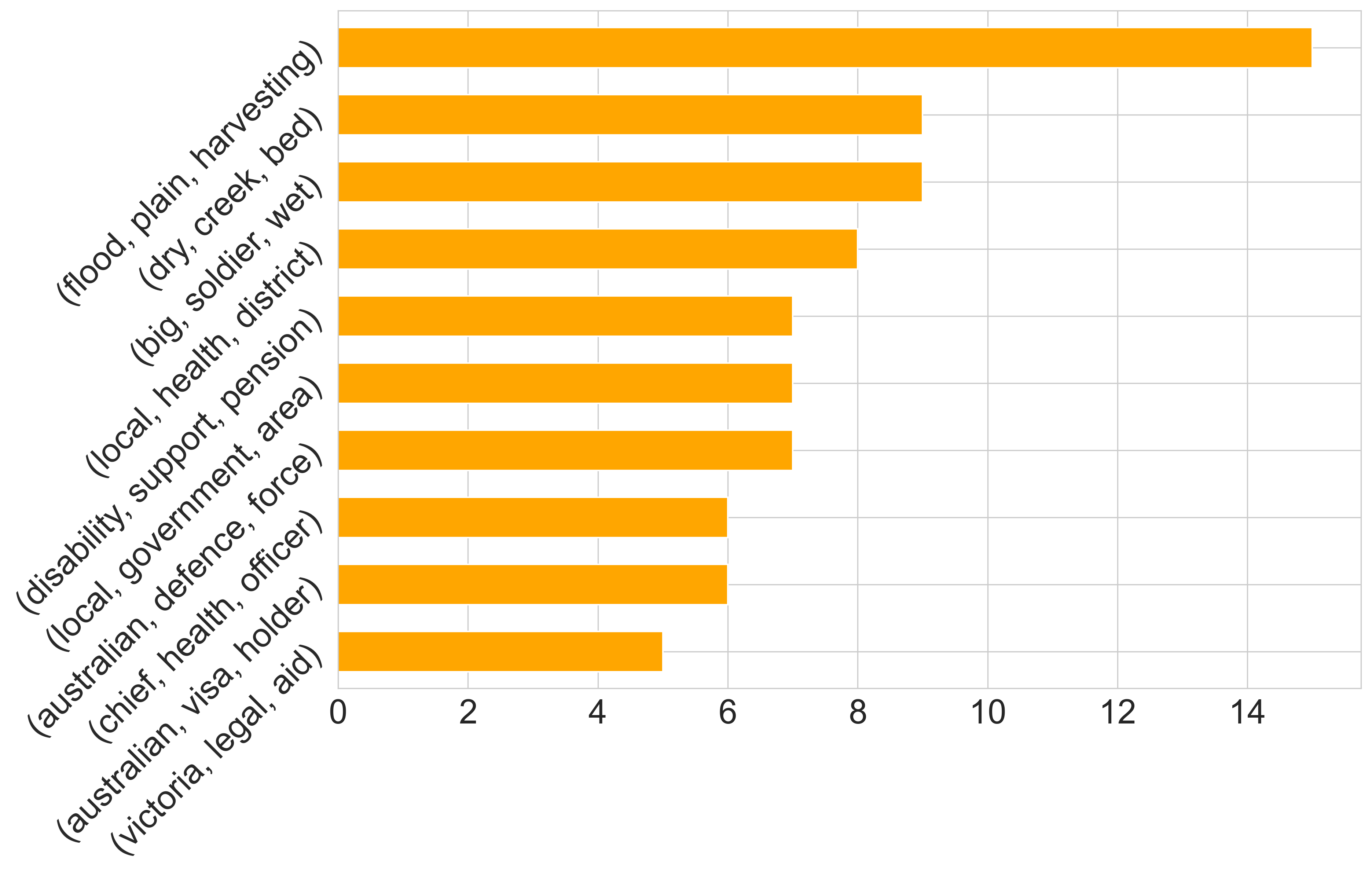}
    \caption{Top 10 trigrams of the Australia news during 2021 third quarter (July - September) for sentiments "sad, annoyed and denial".}
    \label{fig:au21q3}
\end{figure}

 According to Figure \ref{fig:ausdeathcases}, after a small peak in the third quarter 2020, Australia's death toll stabilised until the third quarter 2021, hence we present Figure \ref{fig:au21q3} using the same method, which also contains "sad", "annoyed" and "denial";  we notice that there are no COVID-19 related trigrams except "local health district" and "chief health officer". This may indicate that media and public attention may have shifted to other events, issues or topics, rather than focusing on coverage and discussion of the epidemic. Moreover, this can also indicate the bias in news reporting by The Guardian that gets more attention from grim news articles. 

\begin{figure}[htpb!]
    \centering
    \includegraphics[width=0.45\textwidth]{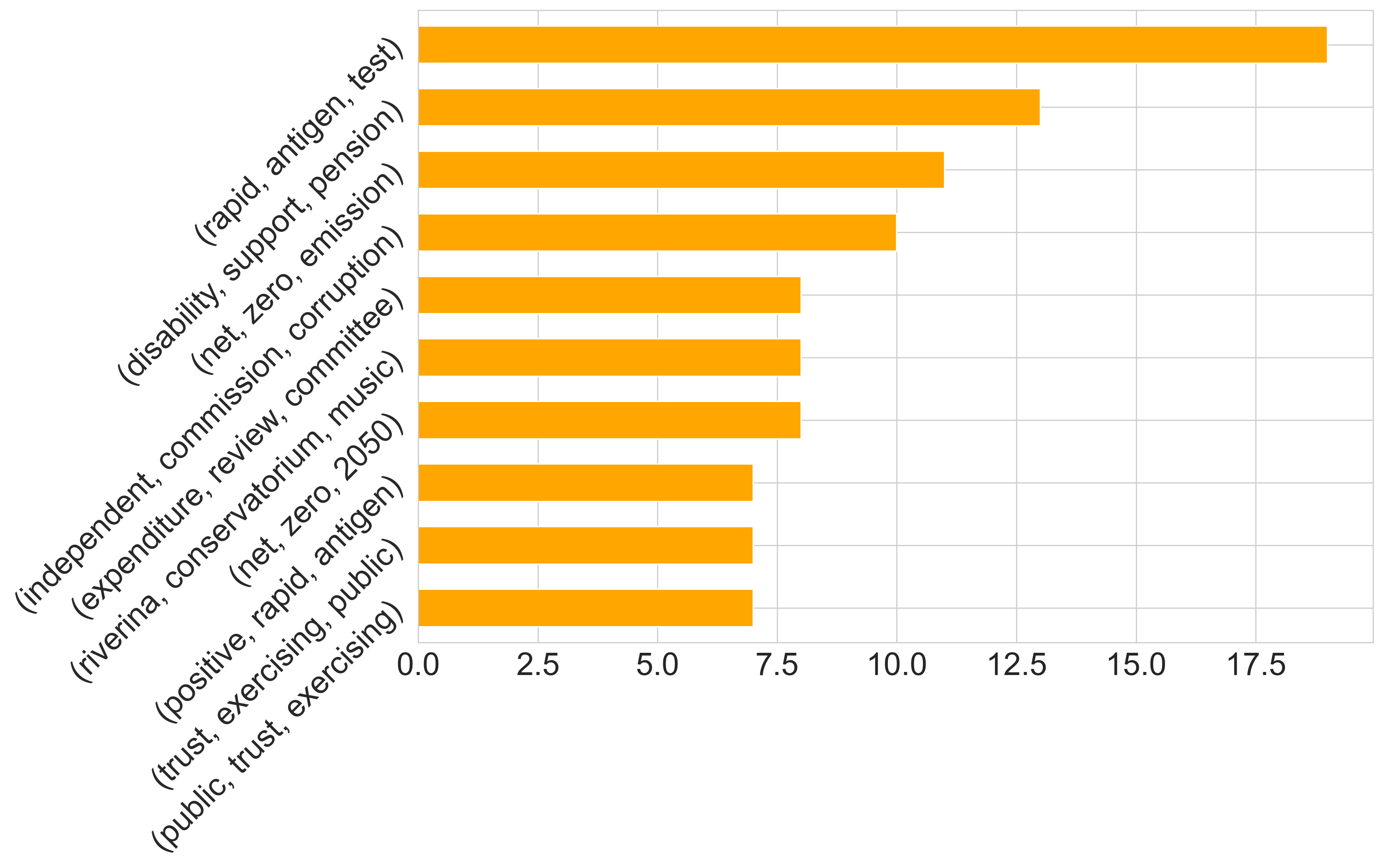} 
    \caption{Top 10 trigrams of the Australia news during 2021 fourth quarter (October - December).}
    \label{fig:au21q4}
\end{figure}

 As the number of deaths continued to rise, the public attention has once again shifted towards COVID-19. Notably, in Figure \ref{fig:au21q4}, the reappearance of the trigrams "rapid antigen test" and "positive rapid antigen" suggests an increase in positive rapid antigen test results. This uptick reflects the escalating number of COVID-19 cases and the accelerated spread of the virus, signalling a renewed focus on the epidemic.

\begin{figure}[htpb!]
    \centering
    \begin{subfigure}{0.45\textwidth}
        \includegraphics[width=\textwidth]{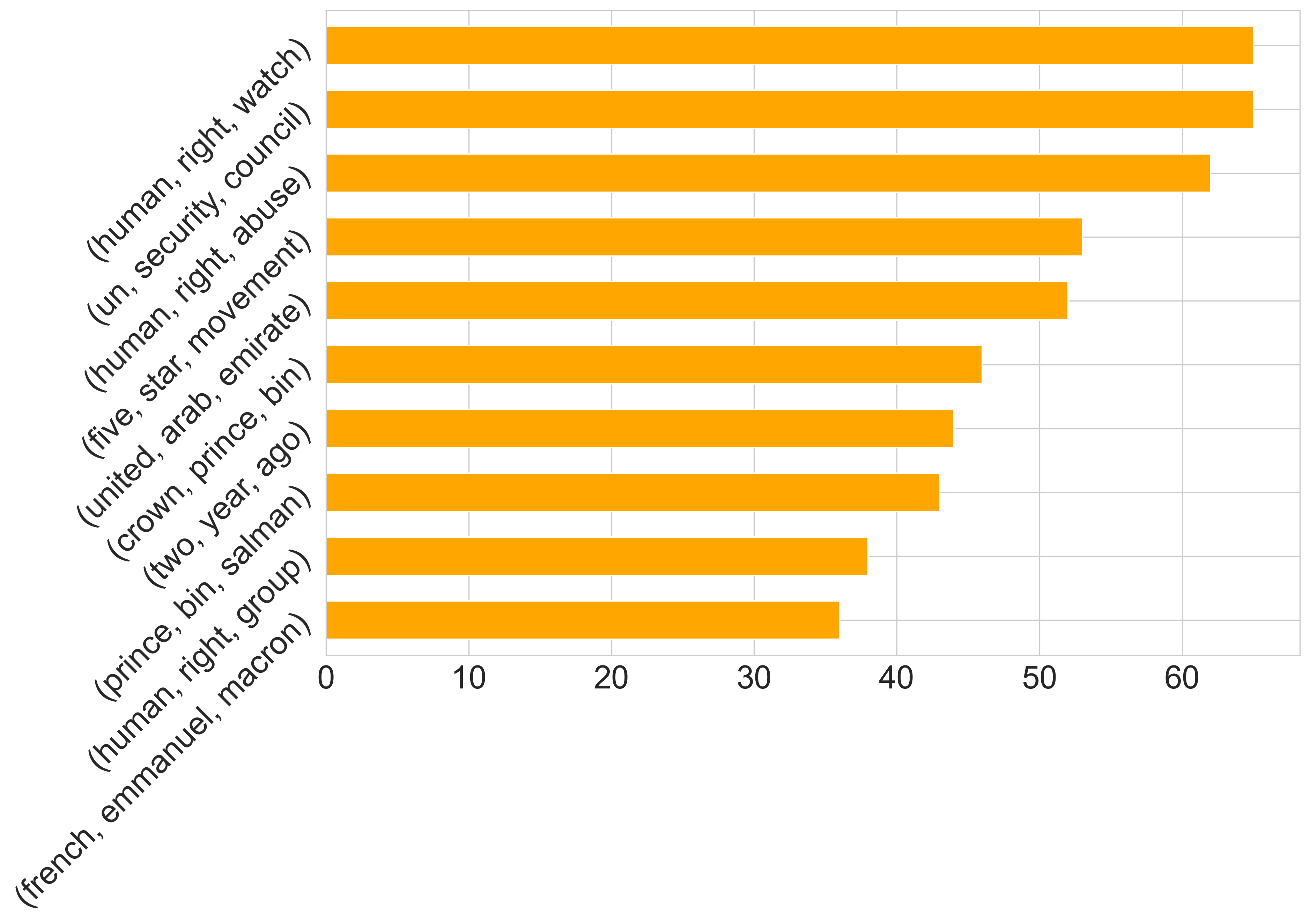}
        \caption{\textcolor{red}{Before COVID-19}}
        \label{fig:triwopre}
    \end{subfigure}
    \begin{subfigure}{0.45\textwidth}
        \includegraphics[width=\textwidth]{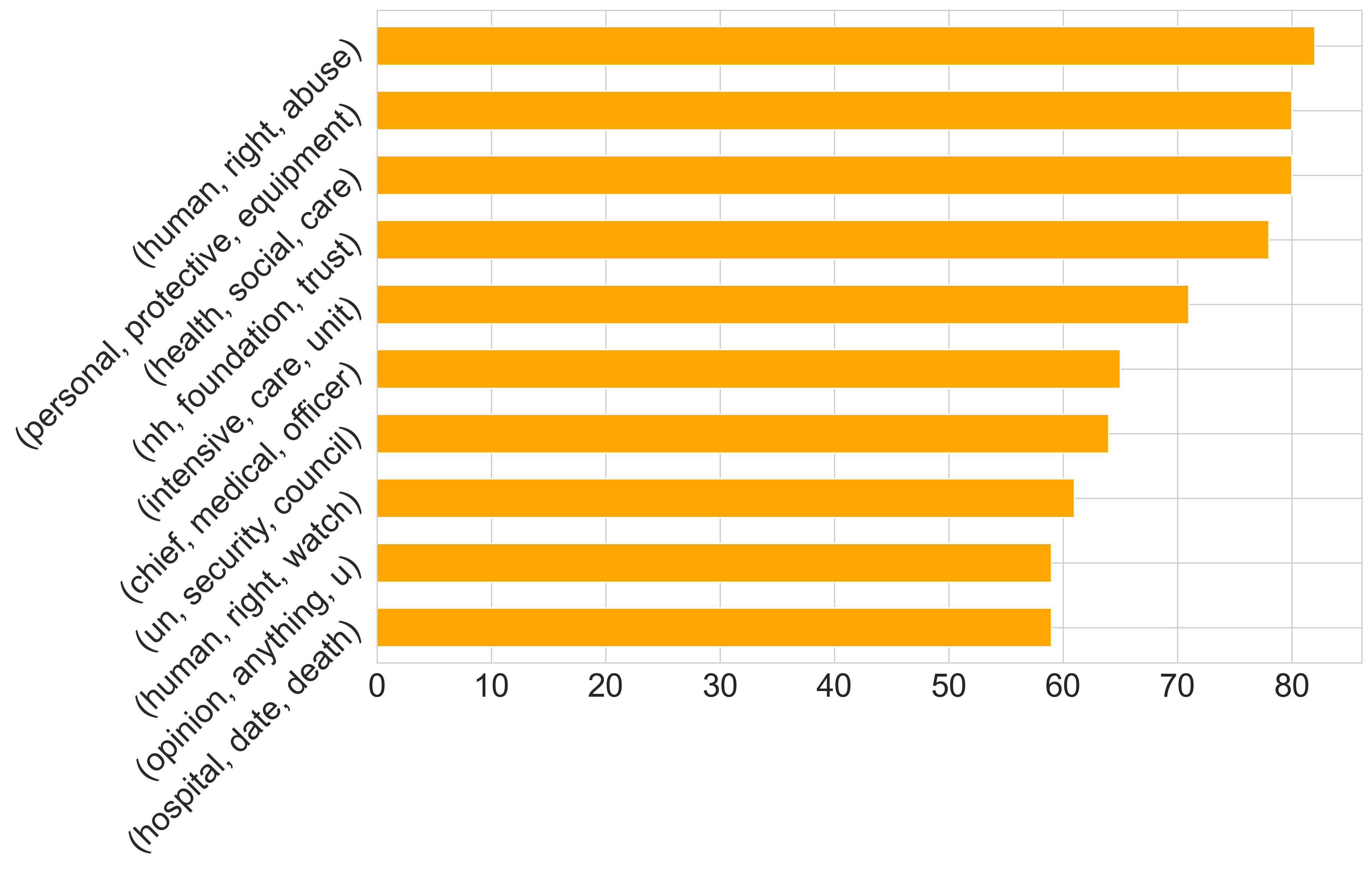}
        \caption{\textcolor{red}{During COVID-19}}
        \label{fig:triwo}
    \end{subfigure}
    \caption{Top 10 trigrams of the world news before and during the COVID-19 pandemic}
    \label{fig:triworldpd}
\end{figure}

 We can see from Figure \ref{fig:triwopre} that before COVID-19, the World News section focused on more diverse articles, ranging from human rights issues to the Prime Minister. Figure \ref{fig:triwo} illustrates the trigrams extracted from The Guardian articles in the World News section during COVID-19. The analysis focused on filtering out sentiments associated with "denial", "annoyed", "anxious", and "sad" from 1st January 2020 to 31st March 2022. Moreover, Figure \ref{fig:triwo}, "health social care", "personal protective equipment" and "intensive care unit (ICU)" were frequently mentioned, reflecting the attention and importance paid to COVID-19 by the world news section, and also demonstrates the concentrated response to the challenges of the epidemic.

\subsection{Examples of sentiments detected}

We selected random samples of Australian news in Table \ref{tab:sample_table3}, which include their predicted sentiments. We can see that in this table,  the classification of "Denial" and "Joking" for the first article is incorrect. There's no indication of denial of reality or humorous elements in the text. Rather, the text discusses the restart of the Titanic II project by Clive Palmer, which doesn't involve denying reality or making jokes. A more appropriate classification for it would be "Optimistic" as it portrays the restart of the project in a positive light, demonstrating hope and positivity for its future success. 

The second row, dated 2021-10-31, describes the experiences of Adnan Choopani and his cousin Mehdi being detained in Australia's immigration detention system for up to eight years. They have witnessed friends self-immolate and have themselves attempted suicide out of despair, all of which are saddening experiences, hence demonstrating sadness. As for pessimism, the text portrays the various hardships and injustices suffered by Adnan and Mehdi in detention, which may evoke a pessimistic outlook on the future, especially considering their prolonged detention despite having lodged protection claims.

The third row indicates anxiety among some unemployed individuals regarding the impending reduction in welfare benefits. They may feel apprehensive and fearful because this change could impact their livelihoods and quality of life.


\begin{table*}[ht]
\centering
\small 
\begin{tabularx}{\textwidth}{|c|X|c|c|}
\hline
\textbf{Sentiments} & \textbf{Sample articles} & \textbf{Public date} & \textbf{Section name} \\
& & & \\
\hline
Denial, Joking &  If the Titanic was the ship that was meant to be unsinkable, the Titanic II is the idea that seems to be un-killable. Six years after Australian mining magnate Clive Palmer declared he was going to build a replica of the Titanic, and three years after he had to suspend work on the project due to money troubles, Palmer has announced that work on the project will start up again. Building the boat – which will have the same interiors and cabin layout as the original vessel, complete with ballroom and Turkish baths, with reports passengers will be given period costumes to wear – has been a long-held dream of Palmer, an Australian businessman and conservative politician...  & 2018-10-24\footnote{\url{https://www.theguardian.com/australia-news/2018/oct/24/titanic-2-australian-billionaire-refloats-dream-clive-palmer}} & Australia news \\
\hline
Pessimistic, Sad & “Time can bring you down,” Adnan Choopani sings, his words echoing off the walls of the detention centre compound, “time can bend your knees”. Time is something Adnan, and his cousin Mehdi, know only too well. For eight years they have been held by Australia’s immigration detention regime, offshore and on. They have watched friends burn themselves to death and known the despair that has led them to attempt suicide themselves. They have been beaten and abused, jailed without reason. They have grown from boys into men in that time. Fifteen and 16 when they arrived in Australia seeking sanctuary, they are now 23. Despite their claims for protection being formally recognised more than half a decade ago, they remain in detention... & 2021-10-31\footnote{\url{https://www.theguardian.com/australia-news/2021/nov/01/time-can-break-your-heart-the-harsh-toll-of-eight-years-in-australian-immigration-detention}} & Australia news \\
\hline
Anxious & Jobseekers in locked-down Melbourne are bracing for a “devastating” \$300 cut to welfare benefits that will hit only two weeks after stage four restrictions are expected to end. The federal government’s plan to taper the coronavirus supplement on 25 September is expected to reduce the incomes of about 2.3 million unemployed people, single parents and students across Australia. But jobseekers and their advocates expect the economic cliff to hit Melbourne particularly hard, with the cut coming only 12 days after the scheduled end to restrictions that shut key industries and imposed a nightly curfew. Cassandra Francisco, 23, who is living under stage 4 restrictions in Footscray in Melbourne’s west, said she was awaiting the change with “absolute dread”.“I’m just going to barely be able to cover my rent,” she said. “I’m already stressed by the situation.”...  & 2020-8-30\footnote{\url{https://www.theguardian.com/australia-news/2020/aug/31/jobseekers-in-locked-down-melbourne-brace-for-devastating-300-cut-to-welfare}} & Australia news \\
\hline
\end{tabularx}
\caption{Sentiment prediction outcomes for randomly selected article samples from Australia sections.}
\label{tab:sample_table3}
\end{table*}

\begin{table*}[ht]
\centering
\small 
\begin{tabularx}{\textwidth}{|c|X|c|c|}
\hline
\textbf{Sentiments} & \textbf{Sample articles} & \textbf{Public date} & \textbf{Section name} \\
& & & \\
\hline
Optimistic, Sad & Monday One of the best family holidays we ever had was on a game reserve in South Africa about 10 years ago. Back then it was normally tricky to get the kids out of bed much before 11, but there were no moans about getting up at six every morning to go on an escorted drive to watch the animals. There was something so magical about being in the presence of such natural beauty. The highlight was coming across three rhinos among a clump of trees. Our jeep crept up to within about 20 metres and we all sat in silence for the best part of an hour, overcome with wonder. So the death of Sudan, the last male northern white rhino, felt more personal than it otherwise might have done... & 2018-03-23\footnote{\url{https://www.theguardian.com/uk-news/2018/mar/23/rhino-death-rotten-kipper-farage-national-humiliation-brexit?CMP=share_btn_url}} & UK news \\
\hline
Annoyed, Joking & Royal tours have long been a central feature of monarchical life. It’s what they do. As the Queen says: “We have to be seen to be believed.” Medieval monarchs toured their realms obsessively in order to show they were still alive. It also helped keep their populations in order and allowed them to display their magnificence and power. Henry II’s legs grew bandy as he rode continuously across France, England and Ireland in the 12th century. Elizabeth I’s tours, 400 years later, wended their way round the country: she spoke to ordinary folk encountered en route and accepted gifts from the burghers of the towns that she and her 300-wagon baggage train passed through... & 2022-03-29\footnote{\url{https://www.theguardian.com/uk-news/2022/mar/29/we-have-to-be-seen-to-be-believed-the-endurance-of-the-royal-tour?CMP=share_btn_url}} & UK news \\
\hline
Annoyed, Denial & Residents, Tory MPs and police have rounded on the government’s handling of new lockdown rules for northern England, while Muslim leaders raised concerns that communities were being scapegoated. On a day of confusion and anger over measures affecting 4.6 million people, police federations warned that new laws barring visitors from private homes or gardens on the eve of Eid may be impossible to enforce. Some MPs expressed anger at the measures in areas with low coronavirus cases, and there was criticism of the announcement being made on Twitter at 9.16pm on Thursday, less than three hours before the rules were imposed. Boris Johnson refused to condemn a fellow Tory MP for saying that Muslims were “just not taking the pandemic seriously”... & 2020-07-31\footnote{\url{https://www.theguardian.com/uk-news/2020/jul/31/matt-hancock-defends-last-minute-northern-coronavirus-lockdown?CMP=share_btn_url}} & UK news \\
\hline
\end{tabularx}
\caption{Sentiment prediction outcomes for randomly selected article samples from UK sections}
\label{tab:sample_table2}
\end{table*}

We show further examples of sentiments detected in selected paragraphs of The Guardian newspaper articles from the UK and World news sections.  We find that the tone of the first row in Table \ref{tab:sample_table2} is mainly optimistic as it tells of a wonderful family holiday experience in a South African game reserve. The author recalls happy and beautiful times spent in nature and emphasizes the positivity of the memories created during that time. The sadness may come from mentioning the death of Sudan, the last male northern white rhino. In the second row, the focus is on providing a historical account of royal tours, highlighting their significance in monarchial life throughout history. Therefore, the tone aligns more closely with an official report rather than expressing annoyance or joking elements. In the third row, the annoyance and denial of some MPs over the new lockdown rules are described. They believe that the implementation of lockdown measures in low-risk areas is unreasonable, and expressed dissatisfaction and criticism that the lockdown regulations were released via Twitter at 9:16 pm on Thursday, less than three hours after the regulations were implemented. Furthermore, Boris Johnson's refusal to condemn a Tory MP's claim that Muslims "just don't take the pandemic seriously" may have caused further negative sentiment.

\begin{table*}[ht]
\centering
\small 
\begin{tabularx}{\textwidth}{|c|X|c|c|}
\hline
\textbf{Sentiments} & \textbf{Sample articles} & \textbf{Public date} & \textbf{Section name} \\
& & & \\
\hline
Annoyed, Joking & Usually, Professor Georgios Babiniotis would take pride in the fact that the Greek word “pandemic” – previously hardly ever uttered – had become the word on everyone’s lips. After all, the term that conjures the scourge of our times offers cast-iron proof of the legacy of Europe’s oldest language. Wholly Greek in derivation – pan means all, demos means people – its usage shot up by more than 57,000\% last year according to Oxford English Dictionary lexicographers. But these days, Greece’s foremost linguist is less mindful of how the language has enriched global vocabulary, and more concerned about the corrosive effects of coronavirus closer to home...  & 2021-01-31\footnote{\url{https://www.theguardian.com/world/2021/jan/31/the-greeks-had-a-word-for-it-until-now-as-language-is-deluged-by-english-terms?CMP=share_btn_url}} & World news \\
\hline
Annoyed, Denial & Simon Tisdall (Trump’s peace plan is a gross betrayal of Afghanistan, Journal, 20 August) is completely correct in his analysis of the US peace initiative in Afghanistan: it is all about enabling the withdrawal of US troops and little about the future of the Afghan people; “details”, such as the shape of the future government of the country, are, apparently, to be settled later. This week saw one historic occasion for Afghanistan, the 100th anniversary of its independence from Britain, but the deal being negotiated in Doha more closely resembles the end of the Soviet engagement in 1989. Desperate to stop a war they were losing, the Soviets negotiated the end of combat operations in 1989 and withdrew the bulk of their troops. Three years later, under pressure from the west, “cooperation troops” withdrew... & 2019-08-23\footnote{\url{https://www.theguardian.com/world/2019/aug/23/afghanistan-at-risk-of-being-abandoned?CMP=share_btn_url}} & World news \\
\hline
Annoyed, Denial & Labour has accused the government of “monumental mistakes” in its handling of the pandemic, as ministers continued to insist they did everything they could to try prevent more than 100,000 deaths from the virus. The shadow health secretary, Jonathan Ashworth, said a “litany of errors” had led to the UK having the fifth-highest death toll in the world and the highest death rate relative to its population. “I just don’t believe that the government did do everything we could,” he told the BBC Radio 4’s Today programme. “We all accept these are challenging times for any government. This is a virus which has swept across the world with speed and severity and it continues to spread ferociously … But monumental mistakes have been made... & 2021-01-27\footnote{\url{https://www.theguardian.com/world/2021/jan/27/monumental-mistakes-made-over-handling-of-covid-says-labour?CMP=share_btn_url}} & World news\\
\hline
\end{tabularx}
\caption{Sentiment prediction outcomes for randomly selected article samples from World news sections}
\label{tab:sample_table1}
\end{table*}

 In Table \ref{tab:sample_table1}, we find that the article discusses Greek scholars' potential pride in the widespread use of the Greek word "pandemic" globally, which may be interpreted as a joke or irony due to the severity of the pandemic with the seemingly pleasant context of using the word. Although the text doesn't directly express anger or annoyance, it suggests that Greek scholars are concerned about the detrimental impact of the epidemic on their homeland, which could manifest as mild annoyance or dissatisfaction. In the second row of Table \ref{tab:sample_table1}, the author expresses annoyance with the United States peace initiative, suggesting that it merely serves as a means to facilitate the withdrawal of their troops without genuinely considering the future and interests of the Afghan people. This dissatisfaction reflects a critical perspective on policy decisions, questioning the sincerity and effectiveness of the initiative. It can be interpreted as a sense of annoyance towards the lack of genuine consideration for the Afghan people's future. Additionally, the author's stance implies a denial of the effectiveness or goodwill of the peace initiative, highlighting scepticism towards its true intentions and outcomes. In the third row of Table \ref{tab:sample_table1}, the Labour Party (opposition)  accused the government of making a "significant mistake" and expressed annoyance, believing that the government's actions were not decisive or effective enough. They pointed out that the United Kingdom had the fifth highest death toll in the world, with the highest mortality rate relative to the population. It also reflects a denial of these measures.

\section{Discussion}

Deep learning models have been prominent in COVID-19 modelling \cite{bhosale2023application} and  sentiment analysis   \cite{alamoodi2021sentiment,chandra2021covid, chandra2023analysis}. In our results, it is evident that the news articles selected from The Guardian during the COVID-19 pandemic mostly contain negative sentiments (denial, sad, annoyed and anxious). In the study conducted by Chandra and Krishna \cite{chandra2021covid}, the emphasis was placed on analysing tweets from specific regions in India, such as Maharashtra and Delhi, during the surge of COVID-19 cases. The results revealed a spectrum of sentiments, predominantly optimism, annoyance, and humour (joking), which were vividly expressed through social media during the pandemic where optimism was the predominant emotion,  but it constitutes a very small proportion in our results. This discrepancy between news media and social media highlights the differing roles that various media types play in public sentiment during crises, suggesting that social media might offer a more diversified emotional reflection, compared to the often grim narrative found in traditional news media.

The differences in emotional expression between social media and traditional news media are pivotal, especially when considering their impact on public perception and behaviour. According to Anspach and Carlson \cite{anspach2020believe}, individuals are more likely to trust and be influenced by the information conveyed through social media comments, even when it deviates from their pre-existing beliefs or is factually incorrect. This tendency to trust social media can be attributed to the perceived immediacy and personal connection users feel towards content creators, which is not as prevalent in traditional media.

The research by Anspach and Carlson \cite{anspach2020believe} demonstrates that despite a general distrust in the news shared through social media, individuals are paradoxically more inclined to believe the information provided by social media posters over content that has been professionally reviewed. This phenomenon can lead to significant implications for public health, where accurate information dissemination is crucial.  Integrating the insights with the findings from our study, which indicate that denial is the most prevalent sentiment, followed by sadness and annoyance, it becomes apparent that the pervasive negative sentiments in COVID-19 news media could have exacerbated a despondent public mood, contrasting with the emotional variety observed on social media. Based on these findings, it is crucial to improve the transparency and accuracy of information dissemination during pandemics to manage public sentiment effectively. Furthermore, we also note that such negative sentiments were also prevalent in The Guardian prior to COVID-19.

\begin{figure}[htpb!]
    \centering
\includegraphics[width=0.5\textwidth]{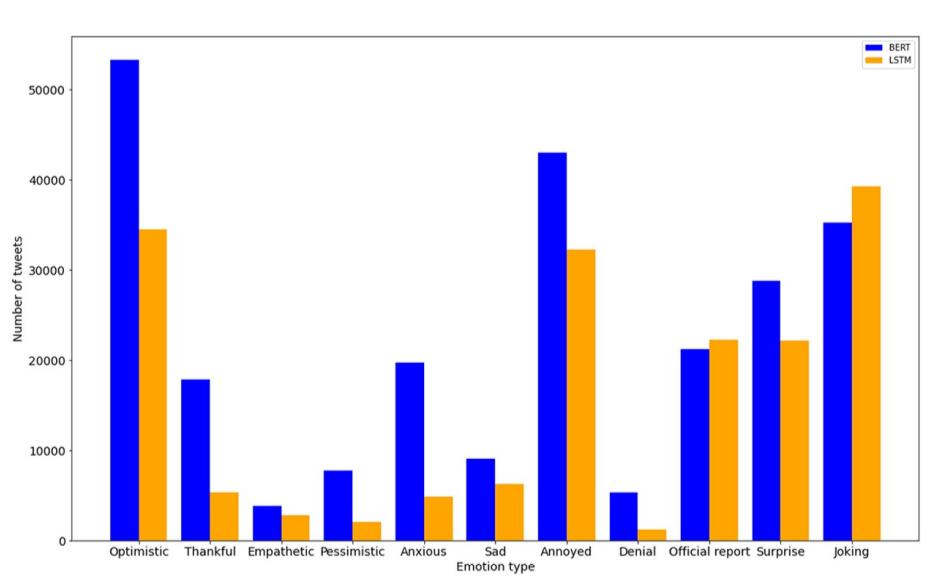}
    \caption{Distribution of sentiments predicted for the Indian tweet dataset by the LSTM and BERT models taken from  Chandra and Krishna (2021)\cite{chandra2021covid}.}
    \label{fig:chandra's_figa}
\end{figure}

Comparing our study with Chandra and Krishna's study in 2021 \cite{chandra2021covid}, the results of sentiment analysis reveal the differences of emotions between social media and traditional media's news reports during the COVID-19 pandemic. According to Fig 5-7 in Chandra and Krishna's 2021 study, positive sentiments such as "optimistic" and "joking" are more prominent in tweets (see Figure\ref{fig:chandra's_figa}). However, from Figure \ref{fig:senBERTvsRoBERTa} in our study, apart from official reports, negative sentiments such as "annoyed" and "denial" are the most prominent. The reason for the sentimental differences among different media may be due to the different sources of text used in sentiment analysis. On social media platforms, everyone could post tweets to share their thoughts about COVID-19, covering sentiments from all users. In traditional media such as The Guardian, news reports were written by newspaper editors, which more represented the official attitude, and needed to use rigorous language to describe various events during the COVID-19 pandemic.

A limitation of this study is that we fine-tuned the model using the SenWave dataset, which consists of millions of tweets and Weibo messages sourced from social media. The information posted on social media often contains numerous abbreviations, internet slang and emojis, which posed a challenge for training our model for sentiment analysis. 
 Furthermore, when training and testing the model, we utilise the tweet dataset, which typically features shorter text lengths when compared to The Guardian articles. As a result, it may fail to capture the same richness and complexity of language features, potentially leading to performance degradation when processing longer texts.
Another limitation is the lack of representativeness in the data we selected. Since the news articles were exclusively sourced from The Guardian, the sentiment analyses  may not reflect the global sentiment towards the COVID-19 pandemic. Additionally, with the increasing prevalence of social media platforms such as Twitter, Instagram, Facebook and Weibo, people are more inclined to share their thoughts and feelings in real-time. Information obtained from news media primarily represents the editorial stance of the outlet, rather than the authentic sentiments of the populace towards the COVID-19 pandemic.

In future work, we can utilise existing models to conduct sentiment analysis on news reports from various news media outlets during the COVID-19 pandemic, aiming to understand the public sentiment in different countries and thereby gain a more comprehensive understanding of the global COVID-19 situation. Meanwhile, more LLMs can be applied to this study, such as the GPT \cite{radford2018improving} model. Building upon the framework of this study, new fine-tuning models can be employed to conduct sentiment analysis on news reports related to various epidemics. Additionally, combining thematic modelling with sentiment analysis, as suggested by Chandra and Krishna \cite{chandra2021covid}, could provide deeper insights into emerging topics relevant to government policies, such as lockdowns and vaccination programs. Furthermore, future research in sentiment analysis can utilise more accurate and effective machine models and related analytical techniques such as multimodal   techniques, integrating various media forms like text, audio, and video.  Specifically, integrating multimodal NLP models into our framework can facilitate the emotional analysis of users' textual, vocal, and facial inputs. This application supports the monitoring and management of individual mental health status and even potentially detects signs of suicidal tendencies or psychological emergencies. These advancements will be applied in such fields to maintain global mental wellness and peace.

\section{Conclusion}

During the COVID-19 pandemic, various media platforms engaged in extensive news reporting, reflecting the complex emotional fluctuations of the public. We  employed large language models, specifically BERT and RoBERTa, to conduct sentiment analysis on news and opinion articles from The Guardian. The results show that both models have similar capabilities when predicting refined sentiments from the SenWave dataset. We proceeded with RoBERTa for further analysis as it provides  advantage over BERT when it comes to computational aspects.

The results indicate that in the initial stages of the pandemic, public sentiment was primarily focused on the urgent response to the crisis. As the situation evolved, the emotional emphasis shifted towards addressing the enduring impacts on health and the economy. Our results report that during the COVID-19 pandemic, the allocation of the emotional tags in news articles from the Guardian indicates a significant increase in the frequency of articles with either one or two emotional labels, highlighting an increased focus on dominant sentiments in news reporting. In comparison of sentiments detected between news articles from Australia and the UK, we found that Australian articles focused more on addressing the immediate crisis, while the UK reports placed greater emphasis on social impacts and mental health. This could be attributed to the UK experiencing a higher severity during the pandemic, which is reflected in the prevalence of negative sentiments in its news reporting, although the reports also frequently exhibited a positive perspective. In comparison with related studies about COVOID-19 sentiment analysis from social media (Tweeter), we found a discrepancy between news media   suggesting that social media offers a more diversified emotional reflection. It is difficult to contest the  grim narrative found in traditional news media such as The Guardian with overall dominance of negative sentiments, pre and during COVID-19.

\section{Code and Data}

 We provide open source code and data that can be used to extend this study:\footnote{\url{https://github.com/sydney-machine-learning/sentimentanalysis-COVID19news}}.

\section*{Acknowledgements} 

We also thank Yiruo Ma and Yifan Dai for their support and contributions to the initial version of this paper. 

\section*{Appendix}

\newpage

\begin{figure*}[htbp]
    \centering
    \begin{subfigure}{0.45\textwidth}
        \includegraphics[width=\linewidth]{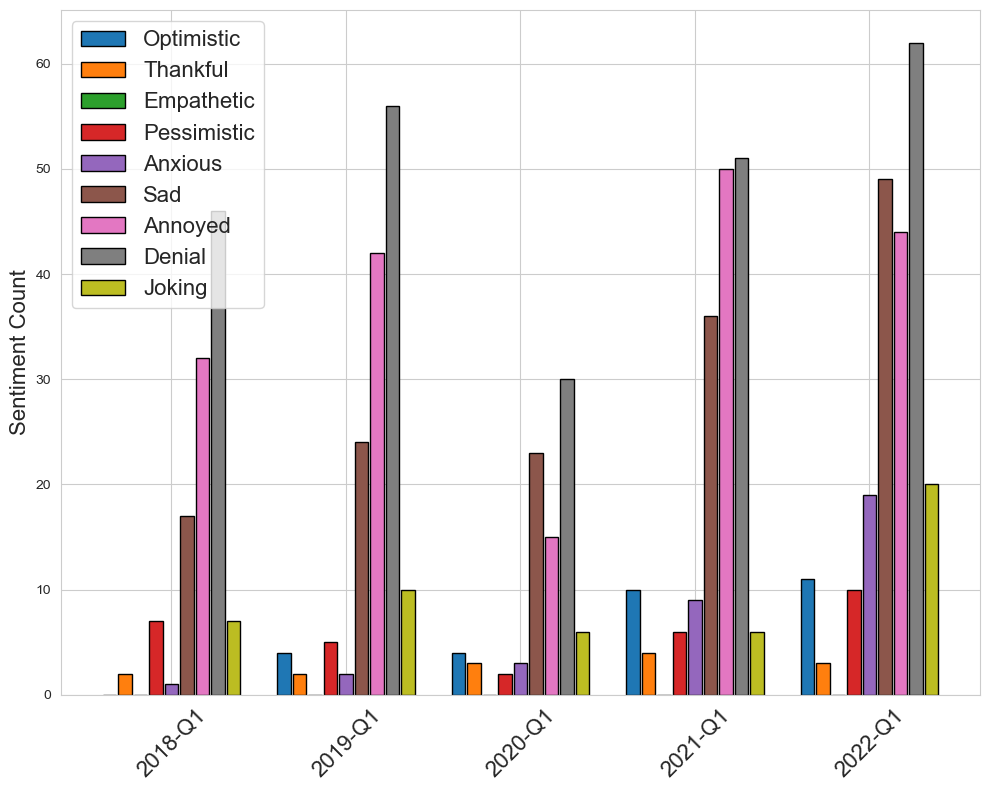}
        \caption{Sentiments detected in the first quarters of years 2018 - 2022}
        \label{fig:Aq1r}
    \end{subfigure}
    \hfill
    \begin{subfigure}{0.45\textwidth}
        \includegraphics[width=\linewidth]{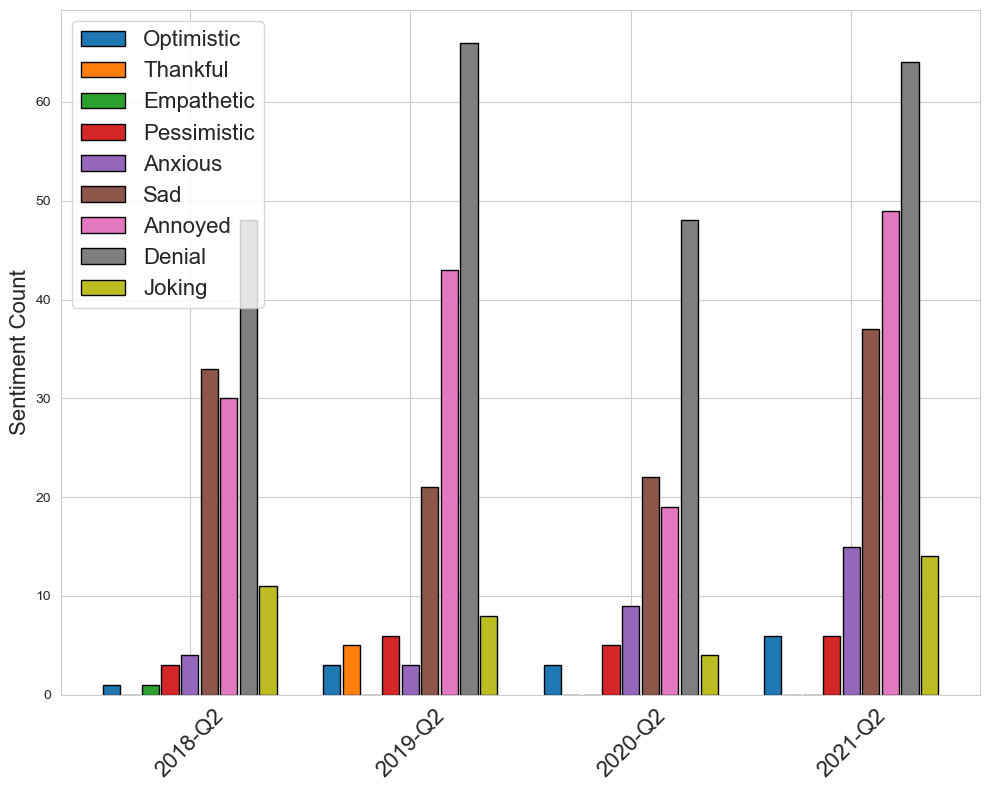}
        \caption{Sentiments detected in the second quarters of years 2018 - 2021}
        \label{fig:Aq2r}
    \end{subfigure}
    \begin{subfigure}{0.45\textwidth}
        \includegraphics[width=\linewidth]{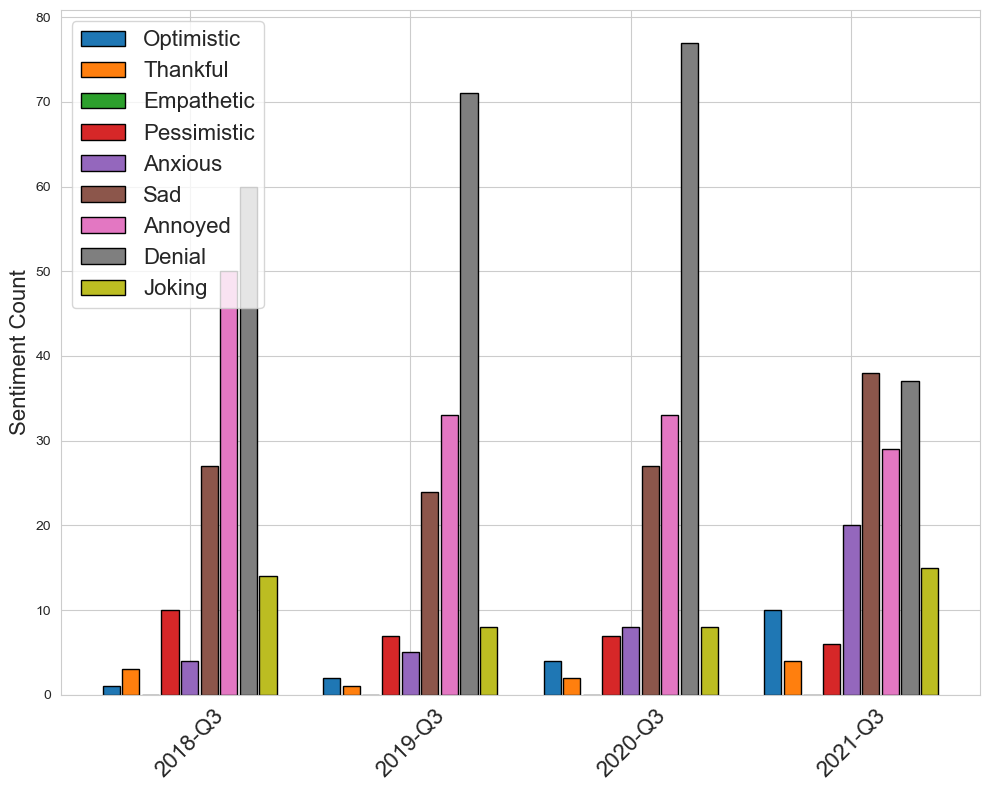}
        \caption{Sentiments detected in the third quarters of years 2018 - 2021}
        \label{fig:Aq3r}
    \end{subfigure}
    \hfill
    \begin{subfigure}{0.45\textwidth}
        \includegraphics[width=\linewidth]{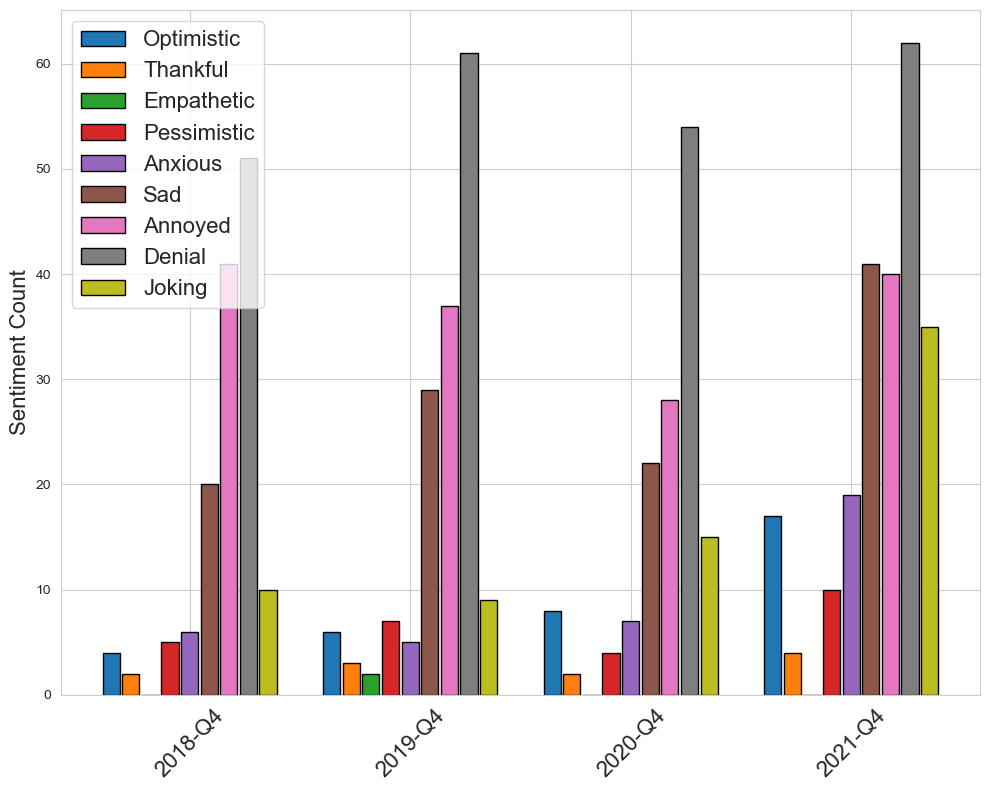}
        \caption{Sentiments detected in the fourth quarters of years 2018 - 2021}
        \label{fig:Aq4r}
    \end{subfigure}
    \caption{Quarterly sentiment counts in Australia news section of The Guardian using the RoBERTa model. }
    \label{fig:quarterly_emotion_disausroberta}
\end{figure*}

In  Figure \ref{fig:quarterly_emotion_disausroberta}, which provides the quarterly sentiment distribution in Australia, we focus on the first quarter of 2022.
The RoBERTa model highlights "sad" as the sentiment with the most significant increase, aligning with earlier observations that RoBERTa is more attuned to identifying sadness. Looking at 2021-Q4 in figure \ref{fig:Aq4r}, the RoBERTa's results are as expected, with fewer instances of "sad" compared to 2022-Q1. The overall trend is clear: there is a general increase in negative sentiment from the pre-pandemic to the pandemic period, underscoring the emotional toll of the COVID-19 crisis as reflected in media coverage across the quarters.

\begin{figure*}[htbp]
    \centering
    \begin{subfigure}{0.45\textwidth}
        \includegraphics[width=\linewidth]{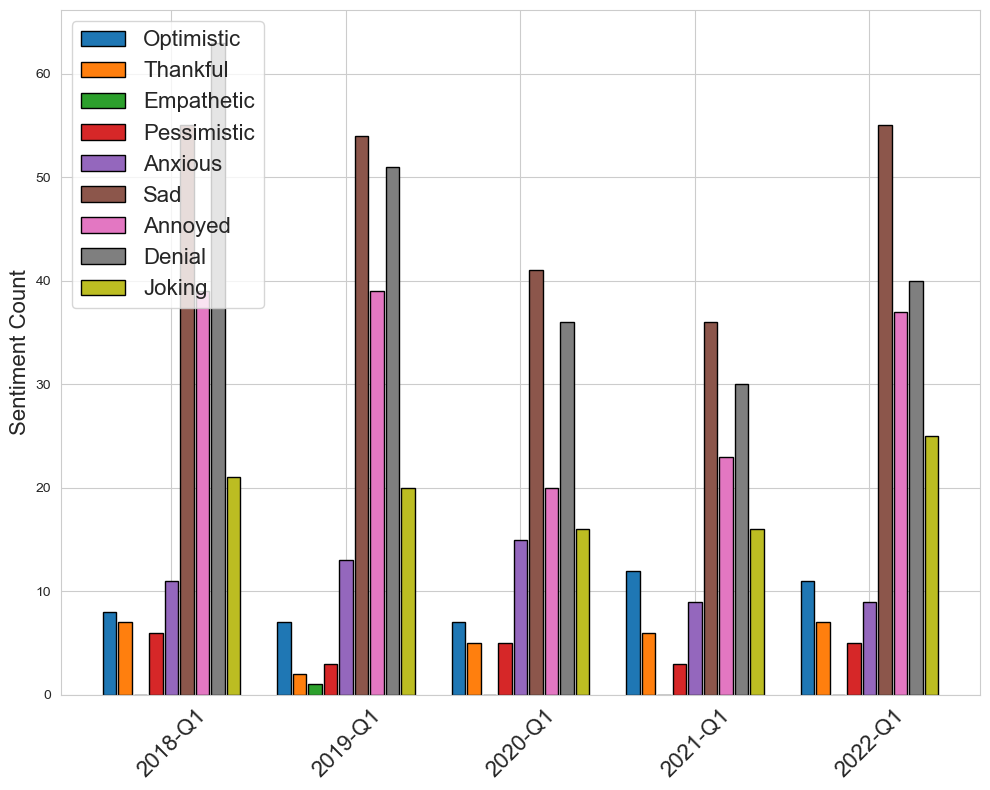}
        \caption{Sentiments detected in the first quarters of years 2018 - 2022}
        \label{fig:uq1r}
    \end{subfigure}
    \hfill
    \begin{subfigure}{0.45\textwidth}
        \includegraphics[width=\linewidth]{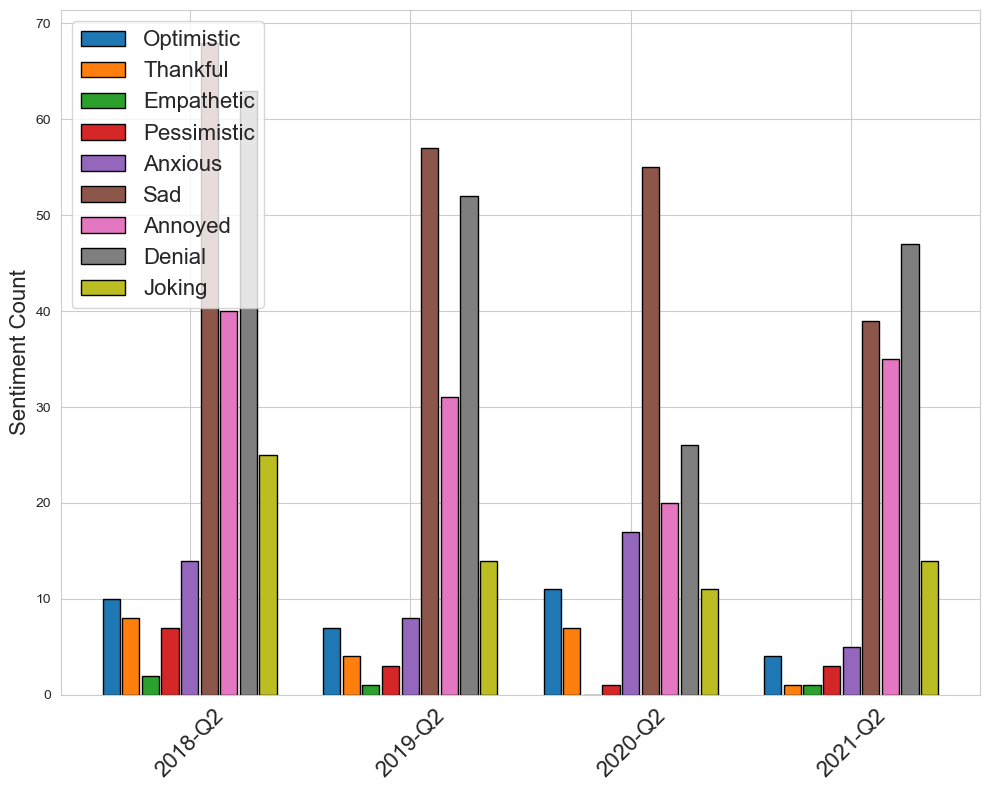}
        \caption{Sentiments detected in the second quarters of years 2018 - 2021}
        \label{fig:uq2r}
    \end{subfigure}

    \begin{subfigure}{0.45\textwidth}
        \includegraphics[width=\linewidth]{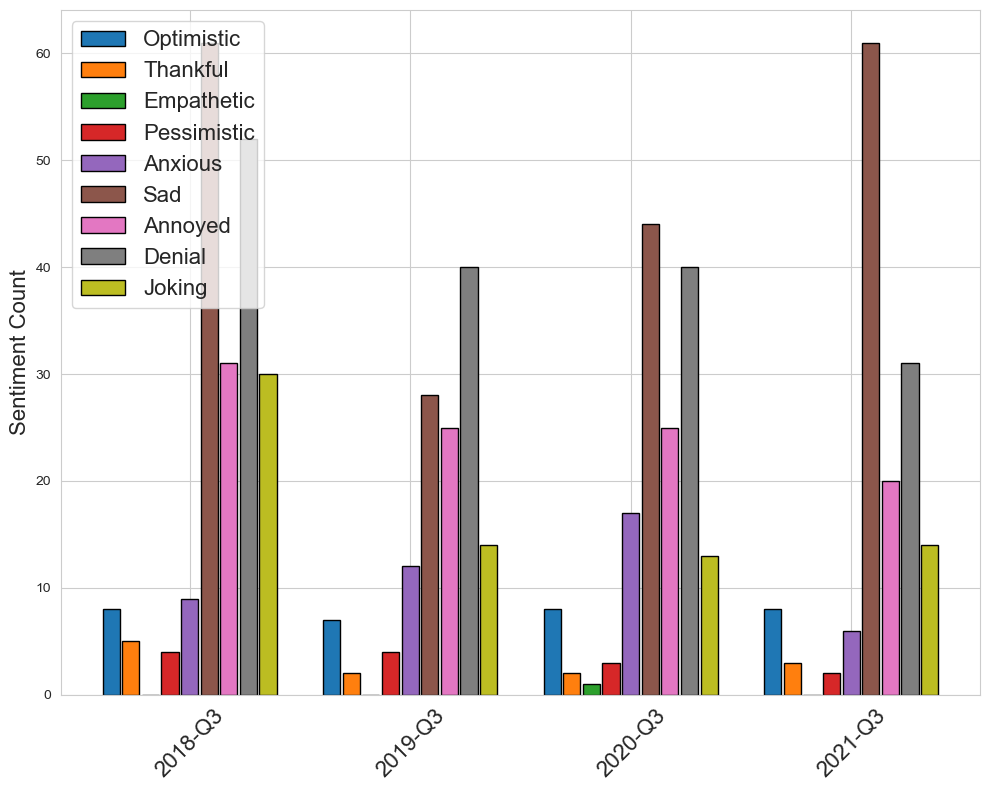}
        \caption{Sentiments detected in the third quarters of years 2018 - 2021}
        \label{fig:uq3r}
    \end{subfigure}
    \hfill
    \begin{subfigure}{0.45\textwidth}
        \includegraphics[width=\linewidth]{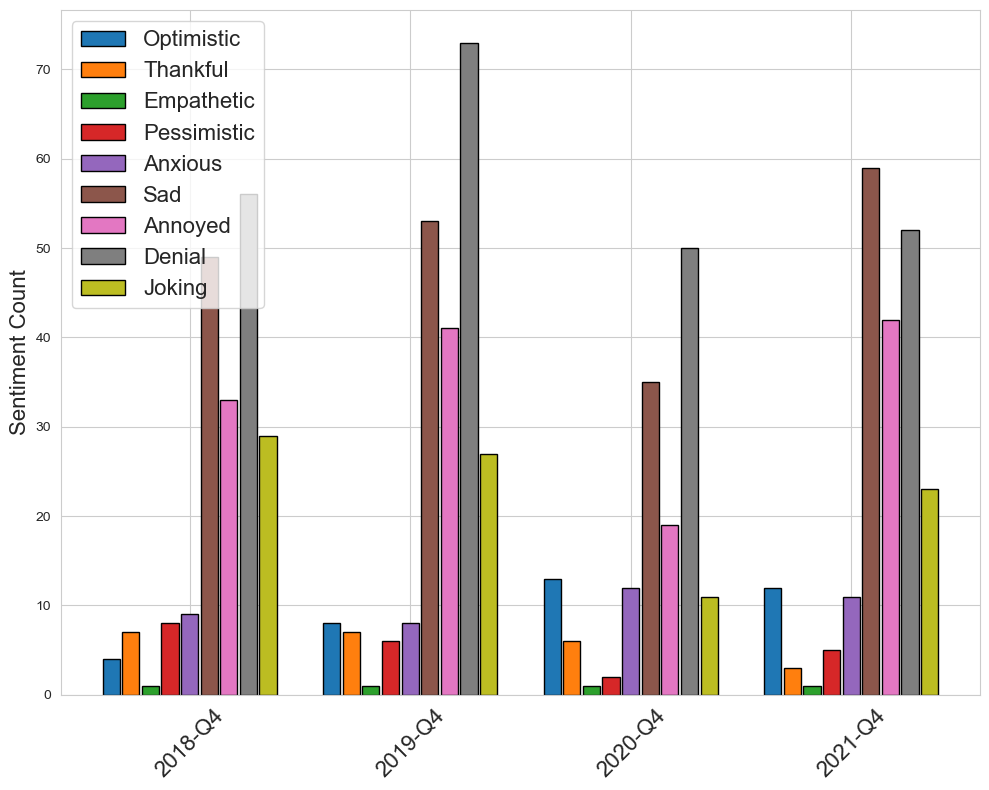}
        \caption{Sentiments detected in the fourth quarters of years 2018 - 2021}
        \label{fig:uq4r}
    \end{subfigure}
    
    \caption{Quarterly sentiment counts in the UK news section of The Guardian using the RoBERTa model. }
    \label{fig:quar_emdisukroberta}
\end{figure*}

Figures  \ref{fig:quar_emdisukroberta}, representing the quarterly sentiment distribution in the UK, contrast with the patterns observed for Australia. According to insights from Figure \ref{fig:ukdeathcases}, the UK's death toll due to COVID-19 exceeded 60,000 in the first quarter of 2021. Interestingly, during this quarter, most negative sentiments like "sad" and "anxious" actually decreased compared to the pre-pandemic period, while positive emotions such as "optimistic" and "thankful" saw a slight increase, especially as identified by the RoBERTa model.

Similarly, despite the death toll reaching over 50,000 in the second quarter of 2020, there was no significant rise in negative emotions detected by RoBERTa. This finding is peculiar when juxtaposed with the Australian data, where a marked increase in negative sentiments correlated with the progression from pre-pandemic to pandemic periods. The emotional distribution in the UK, as reflected in the news, seems to lack a consistent temporal pattern in response to the unfolding crisis

\begin{figure*}[htbp!]
    \centering
    \begin{subfigure}{0.45\textwidth}
        \includegraphics[width=\linewidth]{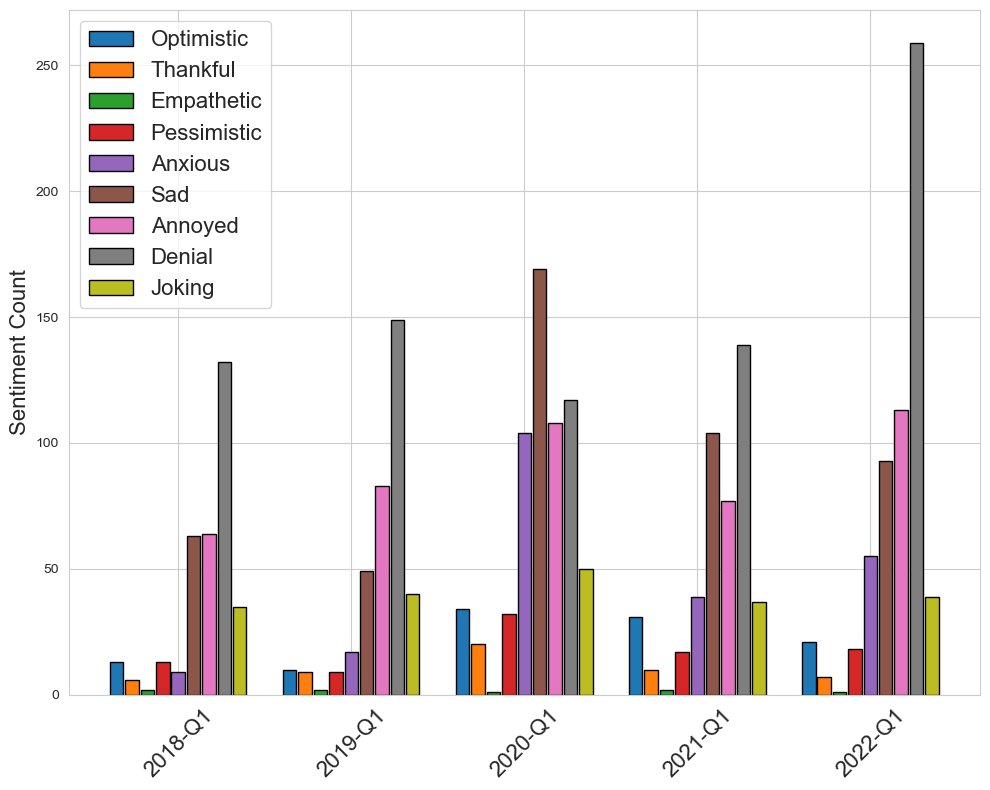}
        \caption{Sentiments detected in the first quarters of years 2018 - 2022}
        \label{fig:wq1r}
    \end{subfigure}
    \hfill
    \begin{subfigure}{0.45\textwidth}
        \includegraphics[width=\linewidth]{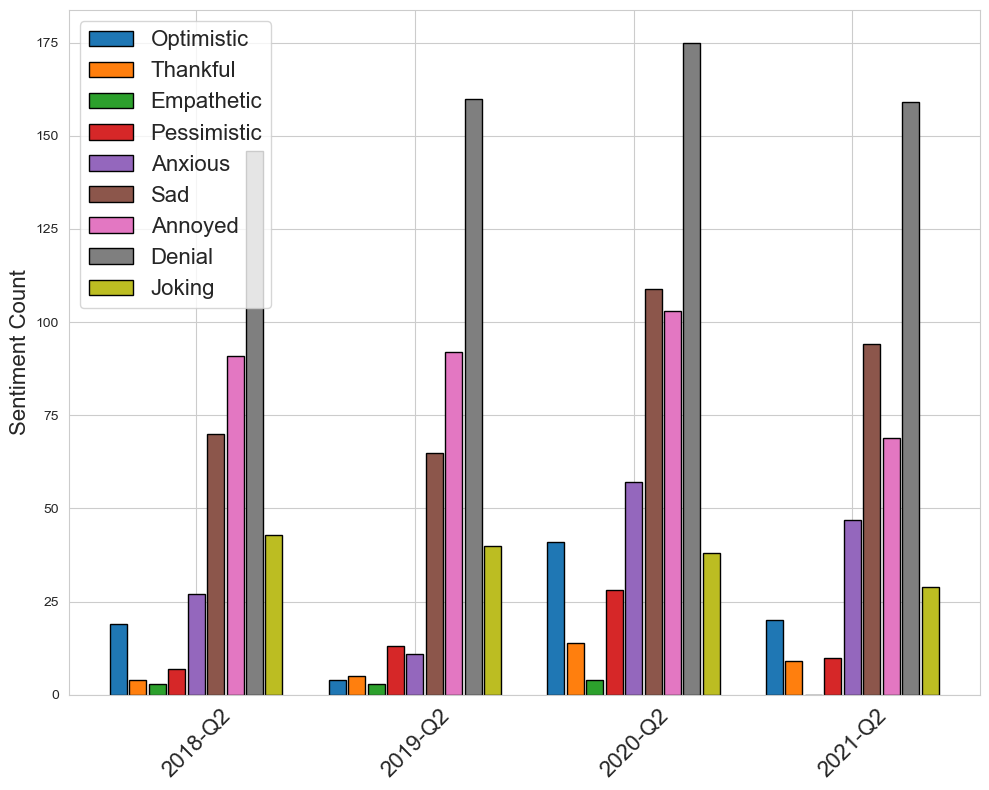}
        \caption{Sentiments detected in the second quarters of years 2018 - 2021}
        \label{fig:wq2r}
    \end{subfigure}

    \begin{subfigure}{0.45\textwidth}
        \includegraphics[width=\linewidth]{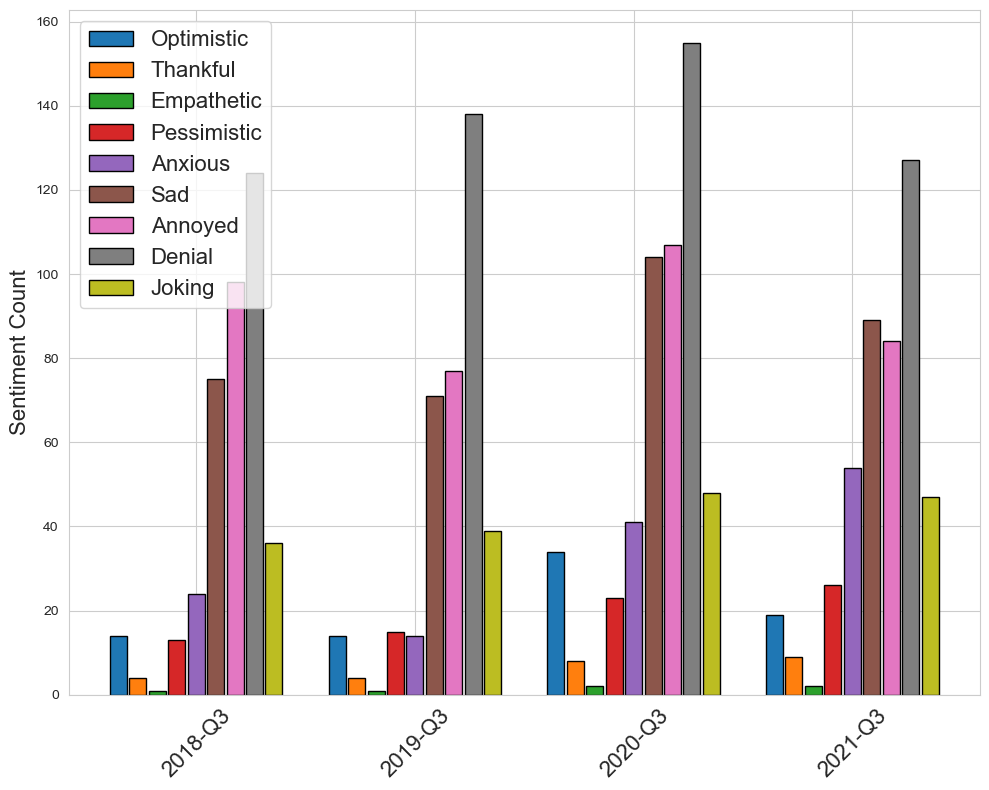}
        \caption{Sentiments detected in the third quarters of years 2018 - 2021}
        \label{fig:wq3r}
    \end{subfigure}
    \hfill
    \begin{subfigure}{0.45\textwidth}
        \includegraphics[width=\linewidth]{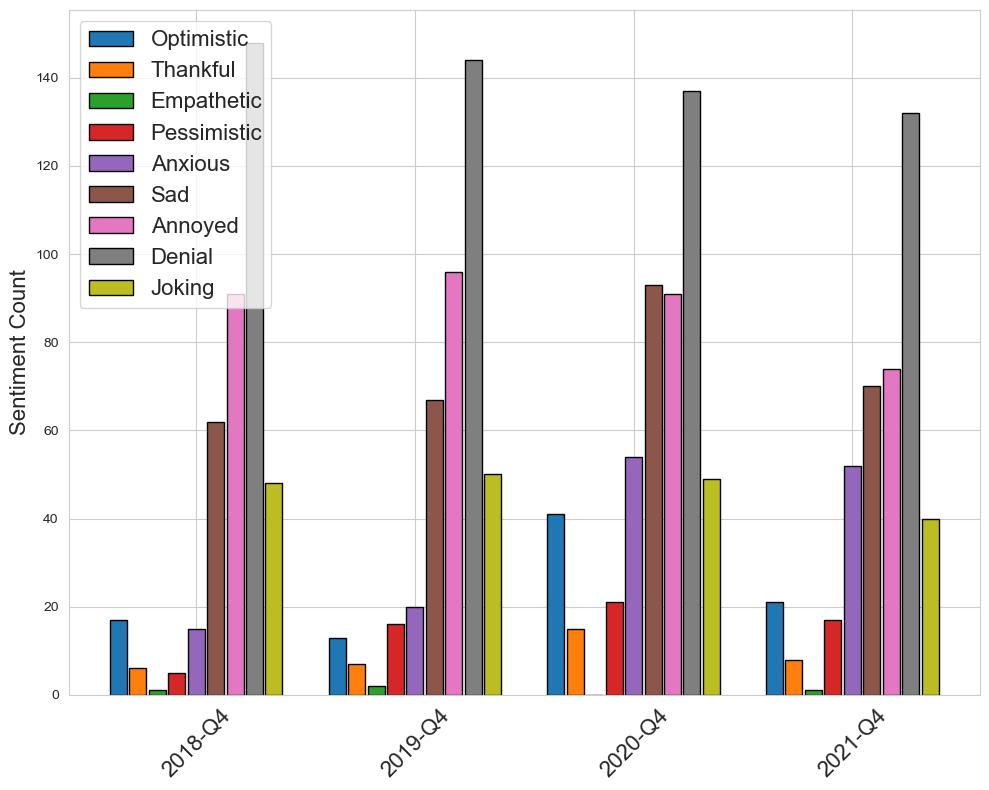}
        \caption{Sentiments detected in the fourth quarters of years 2018 - 2021}
        \label{fig:wq4r}
    \end{subfigure}
    \caption{Quarterly sentiment counts in the world news section of The Guardian using the RoBERTa model. }
    \label{fig:quar_emworldroberta}
\end{figure*}

Figure \ref{fig:quar_emworldroberta} presents  the quarterly emotion distribution worldwide, where RoBERTa models recorded peaks in negative sentiments such as "anxious" and "sad" throughout all four quarters of 2020 — the year WHO officially declared the pandemic \cite{WHOdeclare} —compared to the pre-pandemic years of 2018 and 2019.

\bibliographystyle{elsarticle-num} 
\bibliography{cas-refs}

\begin{thebibliography}{10}
\expandafter\ifx\csname url\endcsname\relax
  \def\url#1{\texttt{#1}}\fi
\expandafter\ifx\csname urlprefix\endcsname\relax\def\urlprefix{URL }\fi
\expandafter\ifx\csname href\endcsname\relax
  \def\href#1#2{#2} \def\path#1{#1}\fi

\bibitem{2020Coronavirus46}
W.~H.~O. (WHO), et~al., Coronavirus disease (covid-19): Situation report-46 (2020).

\bibitem{koh2021deaths}
H.~K. Koh, A.~C. Geller, T.~J. VanderWeele, Deaths from covid-19, Jama 325~(2) (2021) 133--134.

\bibitem{guner2020covid}
H.~R. G{\"u}ner, {\.I}.~Hasano{\u{g}}lu, F.~Akta{\c{s}}, Covid-19: Prevention and control measures in community, Turkish Journal of medical sciences 50~(9) (2020) 571--577.

\bibitem{2023Vaccine}
W.~H.~O. (WHO), et~al., Covid-19 advice for the public: Getting vaccinated (2023).

\bibitem{jin2020virology}
Y.~Jin, H.~Yang, W.~Ji, W.~Wu, S.~Chen, W.~Zhang, G.~Duan, Virology, epidemiology, pathogenesis, and control of covid-19, Viruses 12~(4) (2020) 372.

\bibitem{2021CovidStress}
P.~Klaiber, J.~H. Wen, A.~DeLongis, N.~L. Sin, {The Ups and Downs of Daily Life During COVID-19: Age Differences in Affect, Stress, and Positive Events}, The Journals of Gerontology: Series B 76~(2) (2021) e30--e37.

\bibitem{2021CovidUnrest}
K.~Vandaele, Applauded ‘nightingales’ voicing discontent. exploring labour unrest in health and social care in europe before and since the covid-19 pandemic, Transfer: European Review of Labour and Research 27~(3) (2021) 399--411.

\bibitem{sahni2020role}
H.~Sahni, H.~Sharma, Role of social media during the covid-19 pandemic: Beneficial, destructive, or reconstructive?, International Journal of Academic Medicine 6~(2) (2020) 70--75.

\bibitem{goel2020social}
A.~Goel, L.~Gupta, Social media in the times of covid-19, JCR: Journal of Clinical Rheumatology 26~(6) (2020) 220--223.

\bibitem{chandra2021covid}
R.~Chandra, A.~Krishna, Covid-19 sentiment analysis via deep learning during the rise of novel cases, PloS one 16~(8) (2021) e0255615.

\bibitem{tsao2021social}
S.-F. Tsao, H.~Chen, T.~Tisseverasinghe, Y.~Yang, L.~Li, Z.~A. Butt, What social media told us in the time of covid-19: a scoping review, The Lancet Digital Health 3~(3) (2021) e175--e194.

\bibitem{ahmed2022social}
S.~Ahmed, M.~E. Rasul, Social media news use and covid-19 misinformation engagement: survey study, Journal of Medical Internet Research 24~(9) (2022) e38944.

\bibitem{chowdhary2020natural}
K.~Chowdhary, K.~Chowdhary, Natural language processing, Fundamentals of artificial intelligence (2020) 603--649.

\bibitem{otter2020survey}
D.~W. Otter, J.~R. Medina, J.~K. Kalita, A survey of the usages of deep learning for natural language processing, IEEE transactions on neural networks and learning systems 32~(2) (2020) 604--624.

\bibitem{wu2020deep}
S.~Wu, K.~Roberts, S.~Datta, J.~Du, Z.~Ji, Y.~Si, S.~Soni, Q.~Wang, Q.~Wei, Y.~Xiang, et~al., Deep learning in clinical natural language processing: a methodical review, Journal of the American Medical Informatics Association 27~(3) (2020) 457--470.

\bibitem{li2018deep}
H.~Li, Deep learning for natural language processing: advantages and challenges, National Science Review 5~(1) (2018) 24--26.

\bibitem{jones1994natural}
K.~S. Jones, Natural language processing: a historical review, Current issues in computational linguistics: in honour of Don Walker (1994) 3--16.

\bibitem{brown2020language}
T.~Brown, B.~Mann, N.~Ryder, M.~Subbiah, J.~D. Kaplan, P.~Dhariwal, A.~Neelakantan, P.~Shyam, G.~Sastry, A.~Askell, et~al., Language models are few-shot learners, Advances in neural information processing systems 33 (2020) 1877--1901.

\bibitem{medhat2014sentiment}
W.~Medhat, A.~Hassan, H.~Korashy, Sentiment analysis algorithms and applications: A survey, Ain Shams engineering journal 5~(4) (2014) 1093--1113.

\bibitem{taboada2016sentiment}
M.~Taboada, Sentiment analysis: An overview from linguistics, Annual Review of Linguistics 2 (2016) 325--347.

\bibitem{wankhade2022survey}
M.~Wankhade, A.~C.~S. Rao, C.~Kulkarni, A survey on sentiment analysis methods, applications, and challenges, Artificial Intelligence Review 55~(7) (2022) 5731--5780.

\bibitem{sanchez2020opinion}
P.~S{\'a}nchez-N{\'u}{\~n}ez, M.~J. Cobo, C.~De~Las Heras-Pedrosa, J.~I. Pel{\'a}ez, E.~Herrera-Viedma, Opinion mining, sentiment analysis and emotion understanding in advertising: a bibliometric analysis, IEEE Access 8 (2020) 134563--134576.

\bibitem{shukla2023evaluation}
A.~Shukla, C.~Bansal, S.~Badhe, M.~Ranjan, R.~Chandra, An evaluation of google translate for sanskrit to english translation via sentiment and semantic analysis, Natural Language Processing Journal 4 (2023) 100025.

\bibitem{goodfellow2016deep}
I.~Goodfellow, Y.~Bengio, A.~Courville, Deep learning, MIT press, 2016.

\bibitem{lipton2015critical}
Z.~C. Lipton, J.~Berkowitz, C.~Elkan, A critical review of recurrent neural networks for sequence learning, arXiv preprint arXiv:1506.00019 (2015).

\bibitem{hochreiter1997long}
S.~Hochreiter, J.~Schmidhuber, Long short-term memory, Neural computation 9~(8) (1997) 1735--1780.

\bibitem{yu2019review}
Y.~Yu, X.~Si, C.~Hu, J.~Zhang, A review of recurrent neural networks: {LSTM} cells and network architectures, Neural computation 31~(7) (2019) 1235--1270.

\bibitem{malhotra2016lstm}
P.~Malhotra, A.~Ramakrishnan, G.~Anand, L.~Vig, P.~Agarwal, G.~Shroff, Lstm-based encoder-decoder for multi-sensor anomaly detection, arXiv preprint arXiv:1607.00148 (2016).

\bibitem{vaswani2017attention}
A.~Vaswani, N.~Shazeer, N.~Parmar, J.~Uszkoreit, L.~Jones, A.~N. Gomez, {\L}.~Kaiser, I.~Polosukhin, Attention is all you need, Advances in neural information processing systems 30 (2017).

\bibitem{devlin2019bert}
J.~Devlin, M.-W. Chang, K.~Lee, K.~Toutanova, Bert: Pre-training of deep bidirectional transformers for language understanding (2019).
\newblock \href {http://arxiv.org/abs/1810.04805} {\path{arXiv:1810.04805}}.

\bibitem{acheampong2021transformer}
F.~A. Acheampong, H.~Nunoo-Mensah, W.~Chen, Transformer models for text-based emotion detection: a review of bert-based approaches, Artificial Intelligence Review 54~(8) (2021) 5789--5829.

\bibitem{liu2019roberta}
Y.~Liu, M.~Ott, N.~Goyal, J.~Du, M.~Joshi, D.~Chen, O.~Levy, M.~Lewis, L.~Zettlemoyer, V.~Stoyanov, {RoBERTa}: A robustly optimized {BERT} pretraining approach (2019).
\newblock \href {http://arxiv.org/abs/1907.11692} {\path{arXiv:1907.11692}}.

\bibitem{2022Deep}
D.~Wahyudi, Y.~Sibaroni, Deep learning for multi-aspect sentiment analysis of tiktok app using the rnn-lstm method, Building of Informatics, Technology and Science (BITS) (2022).

\bibitem{kikkisetti2024using}
D.~Kikkisetti, R.~U. Mustafa, W.~Melillo, R.~Corizzo, Z.~Boukouvalas, J.~Gill, N.~Japkowicz, Using llms to discover emerging coded antisemitic hate-speech in extremist social media (2024).
\newblock \href {http://arxiv.org/abs/2401.10841} {\path{arXiv:2401.10841}}.

\bibitem{hart2020politicization}
P.~S. Hart, S.~Chinn, S.~Soroka, Politicization and polarization in covid-19 news coverage, Science communication 42~(5) (2020) 679--697.

\bibitem{basch2020coverage}
C.~H. Basch, A.~Kecojevic, V.~H. Wagner, Coverage of the covid-19 pandemic in the online versions of highly circulated us daily newspapers, Journal of community health 45~(6) (2020) 1089--1097.

\bibitem{yu2021corpus}
H.~Yu, H.~Lu, J.~Hu, A corpus-based critical discourse analysis of news reports on the covid-19 pandemic in china and the uk, International Journal of English Linguistics 11~(2) (2021) 36--45.

\bibitem{kocaman2021spark}
V.~Kocaman, D.~Talby, Spark nlp: natural language understanding at scale, Software Impacts 8 (2021) 100058.

\bibitem{varol2022understanding}
A.~E. Varol, V.~Kocaman, H.~U. Haq, D.~Talby, Understanding covid-19 news coverage using medical nlp, arXiv preprint arXiv:2203.10338 (2022).

\bibitem{evans2023emotional}
S.~L. Evans, R.~Jones, E.~Alkan, J.~S. Sichman, A.~Haque, F.~B.~S. de~Oliveira, D.~Mougouei, The emotional impact of covid-19 news reporting: A longitudinal study using natural language processing, Human behavior and emerging technologies 2023 (2023) 1--16.

\bibitem{tejedor2020information}
S.~Tejedor, L.~Cervi, F.~Tusa, M.~Portales, M.~Zabotina, Information on the covid-19 pandemic in daily newspapers’ front pages: Case study of spain and italy, International journal of environmental research and public health 17~(17) (2020) 6330.

\bibitem{apuke2020nigerian}
O.~D. Apuke, B.~Omar, How do nigerian newspapers report covid-19 pandemic? the implication for awareness and prevention, Health Education Research 35~(5) (2020) 471--480.

\bibitem{sheshadri2017no}
K.~Sheshadri, N.~Ajmeri, J.~Staddon, No (privacy) news is good news: An analysis of new york times and guardian privacy news from 2010--2016, in: 2017 15th Annual Conference on Privacy, Security and Trust (PST), IEEE, 2017, pp. 159--15909.

\bibitem{tunca2023exploratory}
S.~Tunca, B.~Sezen, V.~Wilk, An exploratory content and sentiment analysis of the guardian metaverse articles using leximancer and natural language processing, Journal of Big Data 10~(1) (2023) 82.

\bibitem{mystakidis2022metaverse}
S.~Mystakidis, Metaverse, Encyclopedia 2~(1) (2022) 486--497.

\bibitem{abbasian2017uk}
P.~Abbasian, Uk media coverage of iraq war: A content analysis of tony blair position in the guardian newspaper 2003-2007, Global Media Journal 15~(29) (2017) 77.

\bibitem{Yousefi2021isolation}
R.~Yousefi~Nooraie, K.~Warren, L.~A. Juckett, Q.~A. Cao, A.~C. Bunger, M.~A. Patak-Pietrafesa, Individual- and group-level network-building interventions to address social isolation and loneliness: A scoping review with implications for covid19, PLOS ONE 16~(6) (2021) 1--19.

\bibitem{goel2022investor}
G.~Goel, S.~R. Dash, Investor sentiment and government policy interventions: evidence from covid-19 spread, Journal of Financial Economic Policy 14~(2) (2022) 242--267.

\bibitem{info:doi/10.2196/30765}
M.~Monselise, C.-H. Chang, G.~Ferreira, R.~Yang, C.~C. Yang, Topics and sentiments of public concerns regarding covid-19 vaccines: Social media trend analysis, J Med Internet Res 23~(10) (2021) e30765.

\bibitem{Elman1990}
J.~L. Elman, Finding structure in time, Cognitive science 14~(2) (1990) 179--211.

\bibitem{2014Deep}
J.~Schmidhuber, Deep learning in neural networks: An overview, Neural networks : the official journal of the International Neural Network Society 61 (2014) 85--117.

\bibitem{2015A}
W.~D. Mulder, S.~Bethard, M.~F. Moens, A survey on the application of recurrent neural networks to statistical language modeling, Computer Speech\& Language 30~(1) (2015).

\bibitem{1998The}
S.~Hochreiter, The vanishing gradient problem during learning recurrent neural nets and problem solutions, International Journal of Uncertainty, Fuzziness and Knowledge-Based Systems 06~(2) (1998) --.

\bibitem{2015Critical}
Z.~C. Lipton, J.~Berkowitz, C.~Elkan, A critical review of recurrent neural networks for sequence learning, Computer Science (2015).

\bibitem{2014Dropout}
V.~Pham, T.~Bluche, C.~Kermorvant, J.~Louradour, Dropout improves recurrent neural networks for handwriting recognition, IEEE (2014).

\bibitem{Attention2017}
A.~Vaswani, N.~Shazeer, N.~Parmar, J.~Uszkoreit, L.~Jones, A.~N. Gomez, L.~Kaiser, I.~Polosukhin, Attention is all you need, Advances in neural information processing systems 30 (2017).

\bibitem{BiLSTM2005}
A.~Graves, J.~Schmidhuber, Framewise phoneme classification with bidirectional {LSTM} and other neural network architectures, Neural networks 18~(5-6) (2005) 602--610.

\bibitem{LuWei2022RooB}
D.~Xu, Y.~She, Z.~Tan, R.~Li, J.~Zhao, Research on the recognition of internet buzzword features based on transformer., in: Communications in Computer and Information Science, Vol. 1699, Springer, Singapore, 2022, pp. 227--237.

\bibitem{bansal2023adaptation}
A.~Bansal, A.~Choudhry, A.~Sharma, S.~Susan, Adaptation of domain-specific transformer models with text oversampling for sentiment analysis of social media posts on covid-19 vaccines (2023).
\newblock \href {http://arxiv.org/abs/2209.10966} {\path{arXiv:2209.10966}}.

\bibitem{CunhaWashington2023ACSo}
W.~Cunha, F.~Viegas, C.~França, T.~Rosa, L.~Rocha, M.~A. Gonçalves, A comparative survey of instance selection methods applied to non-neural and transformer-based text classification, ACM computing surveys 55~(13s) (2023) 1--52.

\bibitem{LiGuangju2024MTwm}
G.~Li, D.~Jin, Q.~Yu, Y.~Zheng, M.~Qi, Multiib‐transunet: Transformer with multiple information bottleneck blocks for ct and ultrasound image segmentation, Medical physics (Lancaster) 51~(2) (2024) 1178--1189.

\bibitem{wang2019glue}
A.~Wang, A.~Singh, J.~Michael, F.~Hill, O.~Levy, S.~R. Bowman, {GLUE}: A multi-task benchmark and analysis platform for natural language understanding (2019).
\newblock \href {http://arxiv.org/abs/1804.07461} {\path{arXiv:1804.07461}}.

\bibitem{wang2020superglue}
A.~Wang, Y.~Pruksachatkun, N.~Nangia, A.~Singh, J.~Michael, F.~Hill, O.~Levy, S.~R. Bowman, Super{GLUE}: A stickier benchmark for general-purpose language understanding systems (2020).
\newblock \href {http://arxiv.org/abs/1905.00537} {\path{arXiv:1905.00537}}.

\bibitem{yang2020senwave}
Q.~Yang, H.~Alamro, S.~Albaradei, A.~Salhi, X.~Lv, C.~Ma, M.~Alshehri, I.~Jaber, F.~Tifratene, W.~Wang, T.~Gojobori, C.~M. Duarte, X.~Gao, X.~Zhang, Senwave: Monitoring the global sentiments under the covid-19 pandemic (2020).
\newblock \href {http://arxiv.org/abs/2006.10842} {\path{arXiv:2006.10842}}.

\bibitem{kaggleforecasting}
J.~P.~M. Casper Solheim~Bojer, Kaggle forecasting competitions: An overlooked learning opportunity (2020).
\newblock \href {http://arxiv.org/abs/2009.07701} {\path{arXiv:2009.07701}}.

\bibitem{kaggle_dataset}
A.~KHAROSEKAR, Guardian news articles, \url{https://www.kaggle.com/datasets/adityakharosekar2/guardian-news-articles} (2022).

\bibitem{worldhealthorganization_2021_timeline}
W.~H. Organization, \href{https://www.who.int/emergencies/diseases/novel-coronavirus-2019/interactive-timeline}{Timeline: Who's covid-19 response} (2021).
\newline\urlprefix\url{https://www.who.int/emergencies/diseases/novel-coronavirus-2019/interactive-timeline}

\bibitem{TRIPATHY2016117}
A.~Tripathy, A.~Agrawal, S.~K. Rath, Classification of sentiment reviews using n-gram machine learning approach, Expert Systems with Applications 57 (2016) 117--126.

\bibitem{WuHaiyan2021KnEa}
H.~Wu, Y.~Liu, S.~Shi, Q.~Wu, Y.~Huang, Key n-gram extractions and analyses of different registers based on attention network, Journal of applied mathematics 2021 (2021) 1--16.

\bibitem{Lyse2012}
G.~I. Lyse, G.~Andersen, Collocations and statistical analysis of n-grams: Multiword expressions in newspaper text, Exploring newspaper language: Using the web to create and investigate a large corpus of modern Norwegian, (2012) 79--110.

\bibitem{2011N}
M.~Farhoodi, A.~Yari, A.~Sayah, N-gram based text classification for persian newspaper corpus, in: The 7th international conference on digital content, multimedia technology and its applications, IEEE, 2011, pp. 55--59.

\bibitem{1981Lemmatizing}
W.~Kruase, G.~Willée, Lemmatizing german newspaper texts with the aid of an algorithm, Computers \& the Humanities 15~(2) (1981) 101--113.

\bibitem{2016N}
D.~Nagalavi, M.~Hanumanthappa, N-gram word prediction language models to identify the sequence of article blocks in english e-newspapers, in: International Conference on Computation System \& Information Technology for Sustainable Solutions, 2016, pp. 307--311.

\bibitem{MirandaCarlosHenríquez2023EtEo}
C.~H. Miranda, G.~Sanchez-Torres, D.~Salcedo, Exploring the evolution of sentiment in spanish pandemic tweets: A data analysis based on a fine-tuned bert architecture, Data (Basel) 8~(6) (2023) 96--.

\bibitem{MutindaJames2023SAoT}
J.~Mutinda, W.~Mwangi, G.~Okeyo, Sentiment analysis of text reviews using lexicon-enhanced bert embedding (lebert) model with convolutional neural network, Applied sciences 13~(3) (2023) 1445--.

\bibitem{DangXiaochao2023DRoN}
X.~Dang, L.~Wang, X.~Dong, F.~Li, H.~Deng, Ddert: Research on named entity recognition for mine hoist using a chinese bert model, Electronics (Basel) 12~(19) (2023) 4037--.

\bibitem{loper2002nltk}
E.~Loper, S.~Bird, Nltk: The natural language toolkit, arXiv preprint cs/0205028 (2002).

\bibitem{chhabra2024exploring}
A.~Chhabra, K.~Chaudhary, M.~Alam, Exploring hugging face transformer library impact on sentiment analysis: A case study, in: AI-Based Data Analytics, Auerbach Publications, 2024, pp. 97--106.

\bibitem{pennington2014glove}
J.~Pennington, R.~Socher, C.~D. Manning, Glove: Global vectors for word representation, in: Proceedings of the 2014 conference on empirical methods in natural language processing (EMNLP), 2014, pp. 1532--1543.

\bibitem{huggingface}
T.~Wolf, L.~Debut, V.~Sanh, J.~Chaumond, C.~Delangue, A.~Moi, P.~Cistac, T.~Rault, R.~Louf, M.~Funtowicz, Huggingface's transformers: State-of-the-art natural language processing (2020).
\newblock \href {http://arxiv.org/abs/1910.03771} {\path{arXiv:1910.03771}}.

\bibitem{a2023_covid19}
ABS, \href{https://www.abs.gov.au/articles/covid-19-mortality-australia-deaths-registered-until-30-september-2023}{Covid-19 mortality in australia: Deaths registered until 30 september 2023 | australian bureau of statistics} (2023).
\newline\urlprefix\url{https://www.abs.gov.au/articles/covid-19-mortality-australia-deaths-registered-until-30-september-2023}

\bibitem{ons_2022_deaths}
ONS, \href{https://www.ons.gov.uk/peoplepopulationandcommunity/birthsdeathsandmarriages/deaths/datasets/weeklyprovisionalfiguresondeathsregisteredinenglandandwales}{Deaths registered weekly in england and wales, provisional - office for national statistics} (2022).
\newline\urlprefix\url{https://www.ons.gov.uk/peoplepopulationandcommunity/birthsdeathsandmarriages/deaths/datasets/weeklyprovisionalfiguresondeathsregisteredinenglandandwales}

\bibitem{tsoumakas2007multi}
G.~Tsoumakas, I.~Katakis, Multi-label classification: An overview, International Journal of Data Warehousing and Mining (IJDWM) 3~(3) (2007) 1--13.

\bibitem{a2020_the}
\href{https://www.theguardian.com/info/ng-interactive/2020/dec/21/the-guardian-in-2020}{The guardian in 2020} (2020).
\newline\urlprefix\url{https://www.theguardian.com/info/ng-interactive/2020/dec/21/the-guardian-in-2020}

\bibitem{bhosale2023application}
Y.~H. Bhosale, K.~S. Patnaik, Application of deep learning techniques in diagnosis of {COVID-19 (coronavirus)}: a systematic review, Neural processing letters 55~(3) (2023) 3551--3603.

\bibitem{alamoodi2021sentiment}
A.~H. Alamoodi, B.~B. Zaidan, A.~A. Zaidan, O.~S. Albahri, K.~I. Mohammed, R.~Q. Malik, E.~M. Almahdi, M.~A. Chyad, Z.~Tareq, A.~S. Albahri, et~al., Sentiment analysis and its applications in fighting {COVID-19} and infectious diseases: A systematic review, Expert systems with applications 167 (2021) 114155.

\bibitem{chandra2023analysis}
R.~Chandra, J.~Sonawane, J.~Lande, C.~Yu, An analysis of vaccine-related sentiments from development to deployment of covid-19 vaccines (2023).
\newblock \href {http://arxiv.org/abs/2306.13797} {\path{arXiv:2306.13797}}.

\bibitem{anspach2020believe}
N.~M. Anspach, T.~N. Carlson, What to believe? social media commentary and belief in misinformation, Political Behavior 42~(3) (2020) 697--718.

\bibitem{radford2018improving}
A.~Radford, K.~Narasimhan, T.~Salimans, I.~Sutskever, et~al., Improving language understanding by generative pre-training (2018).

\bibitem{WHOdeclare}
W.~H.~O. (WHO), et~al., Who director-general's opening remarks at the media briefing on covid-19 - 11 march 2020 (2020).

\end{thebibliography}





\end{document}